\documentclass[12pt]{l4dc2023}

\title[Compositional Learning-based Planning for Vision POMDPs]{Compositional Learning-based Planning for Vision POMDPs}
\usepackage{times}

\usepackage{graphicx} 
\usepackage{mathptmx} 
\usepackage{amsmath} 
\usepackage{amssymb}  
\usepackage[capitalize]{cleveref}
\usepackage{booktabs}
\usepackage{xcolor}
\usepackage{colortbl}
\allowdisplaybreaks[4]
\usepackage{algorithm}
\usepackage{algorithmic}
\usepackage{wrapfig}
\usepackage[font=small]{caption}

\usepackage{todonotes}
\definecolor{Gray}{gray}{0.975}
\newcolumntype{a}{>{\columncolor{Gray}}l}

\author{%
 \Name{Sampada Deglurkar}$^{*,1}$ \Email{sampada\_deglurkar@berkeley.edu}\\
 \Name{Michael H. Lim}$^{*,1}$ \Email{michaelhlim@berkeley.edu}\\
 \Name{Johnathan Tucker}$^{2}$ \Email{jotu9759@colorado.edu}\\
 \Name{Zachary N. Sunberg}$^{2}$ \Email{zachary.sunberg@colorado.edu}\\
 \Name{Aleksandra Faust}$^{3}$ \Email{faust@google.com}\\
 \Name{Claire J. Tomlin}$^{1}$ \Email{tomlin@eecs.berkeley.edu}\\
 \addr $^{1}$UC Berkeley, $^{2}$CU Boulder, $^{3}$Google Research.
}

\begin{document}

\maketitle

\begin{abstract}
  The Partially Observable Markov Decision Process (POMDP) is a powerful framework for capturing decision-making problems that involve state and transition uncertainty. 
  However, most current POMDP planners cannot effectively handle high-dimensional image observations prevalent in real world applications, and often require lengthy online training that requires interaction with the environment.
  In this work, we propose Visual Tree Search (VTS), a compositional learning and planning procedure that combines generative models learned offline with online model-based POMDP planning.
  The deep generative observation models evaluate the likelihood of and predict future image observations in a Monte Carlo tree search planner.
  We show that VTS is robust to different types of image noises that were not present during training and can adapt to different reward structures without the need to re-train.
  This new approach significantly and stably outperforms several baseline state-of-the-art vision POMDP algorithms while using a fraction of the training time.
\end{abstract}


\section{Introduction}\label{sec:intro}
Many sequential decision making problems, such as autonomous driving \citep{bai2015intention,sunberg2017value}, cancer screening \citep{ayer2012mammography}, spoken dialog systems \citep{young2013pomdp}, and aircraft collision avoidance~\citep{holand2013optimizing}, involve uncertainty in both sensing and planning.
Planning under partial observability is challenging, as the agent must address both localization and uncertainty-aware planning through active information gathering in the environment.
By capturing the observation uncertainty through a \textit{belief} distribution over possible states, the agent will be able to fully close the observation-plan-action loop.
While there are many methods that can either handle visual localization~\citep{Jonschkowski-RSS-2018,karkus2020dmnet,Karkus2021slamnet} or planning under uncertainty~\citep{van2012motion,todorov2005ilqg}, a na\"ive combination of these methods, for instance by assuming the certainty equivalence principle or simplifying the observation space, may not yield meaningful closed-loop control policies that enable active information gathering under more general environmental assumptions.

The partially observable Markov decision process (POMDP) formalism is a powerful framework that can capture and systematically solve these sequential decision making under uncertainty problems.
However, finding an optimal POMDP policy is computationally demanding, and often intractable, due to the uncertainty introduced by imperfect observations~\citep{papadimitriou1987complexity}.
One popular approach to deal with this challenge is to use \emph{online} algorithms that look for local approximate policies as the agent interacts with the environment rather than a global policy that maps every possible outcome to an action, such as Monte Carlo tree search (MCTS) and similar variants~\citep{browne2012,Silver2010,Sunberg2017,ye2017despot,kurniawati2016online}.
Many of these state-of-the-art MCTS algorithms enjoy computational efficiency \citep{Sunberg2017,mern2020,lim2021voronoi} and finite sample convergence guarantees to the optimal policy \citep{Lim2020,lim2021voronoi}.
Despite their flexibility and optimality, these methods rely on having access to generative models and observation density models, which limits the class of problems they can solve in practice.
In many realistic scenarios with high dimensional observations like RGB images, these POMDP methods cannot be applied without knowing or learning the relevant models or simplifying the environment.

Recently, there has been an increased interest in solving \textit{vision POMDPs}, i.e. POMDPS with image or video observations, using deep learning methods.
Model-free vision POMDP algorithms train an end-to-end deep neural network policy to learn both a latent belief representation and a planner \citep{Karkus2017,mnih-atari-2013,igl2018dvrl}, which benefit from not having to specify the transition and observation models and can learn complex policies.
However, they may lack interpretability, not generalize well to new unseen tasks, and not leverage much prior knowledge about the system, especially in robotics settings.
In contrast, model-based vision POMDP algorithms~\citep{wang2020,singh21worldbelief} combine classical filtering and planning techniques with deep learning.
While such algorithmic structure allows the models to focus on specific tasks, making them sample efficient and robust, these methods often rely on simplified approaches for planning in the belief space and require learning an advantage function that only partially captures uncertainty by coupling observations with rewards.

\begin{figure}[t]
  \centering
  \includegraphics[width=0.8\textwidth]{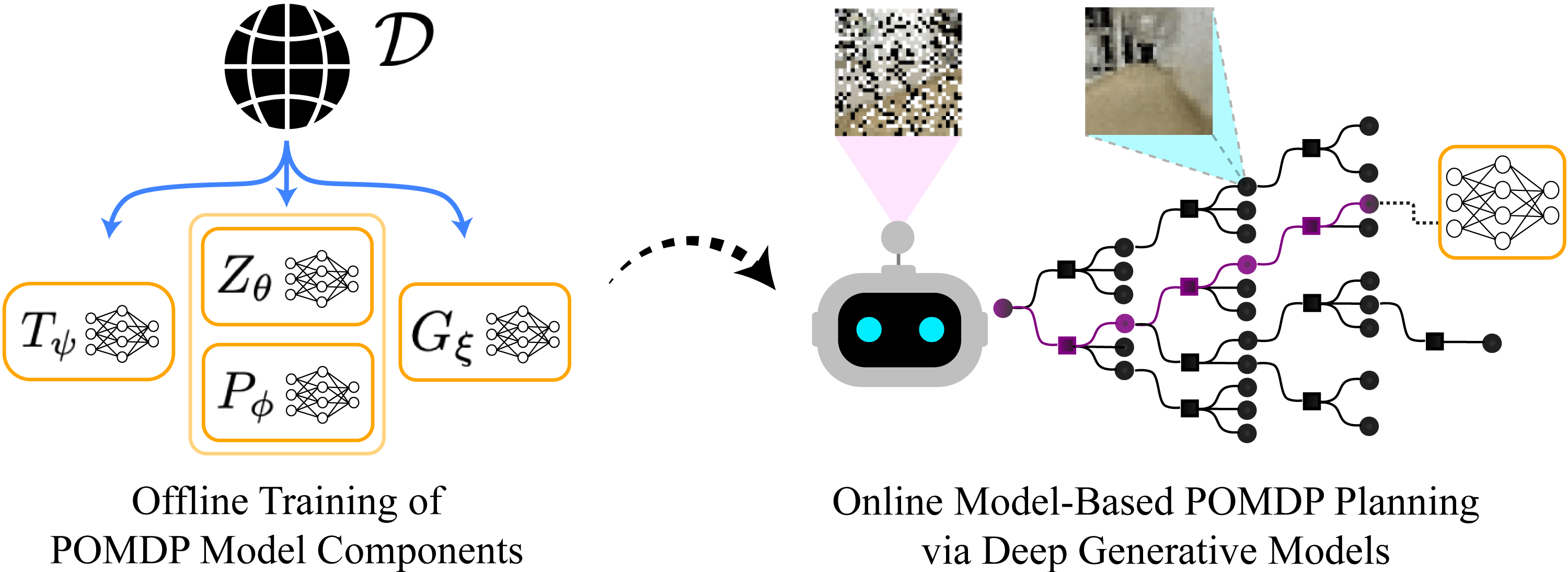}
  \caption{Visual Tree Search interfaces offline deep generative model training with online model-based POMDP filtering and planning.}
  \label{fig:vts-overview}
\end{figure}

Thus, we propose Visual Tree Search (VTS), a procedure to solve vision POMDPs by combining deep generative models and online tree search planning, effectively framing the POMDP reinforcement learning problem as a compositional unsupervised learning problem. 
Our key insight is to utilize compositional learning approaches to bridge the offline training of individual sets of models with online model-based planning that uses learning-enabled components.
Introducing the algorithmic prior knowledge of particle filtering and MCTS decreases the computational complexity required by the learning and  
captures state and transition uncertainty without dependence on reward structure.
Our empirical analyses demonstrate that tackling uncertainty via VTS enhances performance, robustness, and interpretability of neural network components.


\section{Background}\label{sec:background}
\paragraph{POMDPs.} A POMDP is defined with a 7-tuple $(S,A,O,T,Z,R,\gamma)$, with state space $S$, action space $A$, observation space $O$, transition density $T(s'|s,a)$, observation density $Z(o|s)$, reward function $R(s,a)$, and discount $\gamma \in [0,1)$ \citep{kochenderfer2015decision,bertsekas2005dynamic}. 
Specifically, we say ``continuous POMDPs'' to denote those with continuous state, action, and observation spaces.
For POMDPs, since the agent receives only noisy observations of the true state, it can infer the state by maintaining a belief $b_t \in B$ at each step $t$ and updating it with the new action and observation pair $(a_{t+1},o_{t+1})$ via Bayesian filtering \citep{kaelbling1998planning}. 
A policy $\pi:B\to A$ maps a belief $b$ to an action $a$. 
The agent seeks to find an optimal policy $\pi^*$ that maximizes the expected cumulative reward.

\paragraph{Monte Carlo Tree Search.}
In Monte Carlo planning, it is not always necessary to evaluate the exact probability of all transitions and observations, and merely generating samples of next state $s'$, reward $r$, and observation $o$ is sufficient. 
In particular, many Monte Carlo Tree Search (MCTS) algorithms only require that we have generative models that can generate $(s',r,o)$ samples with densities $s' \sim T(s'|s,a)$ and $o\sim G(o|s')$, and observation density models that can evaluate the likelihood $w = Z(o|s')$.
Using these samples, MCTS can reason about the transition and observation densities by balancing exploration and exploitation to approximate the true reward distribution. 
This results in an approximate online policy that maximizes the expected sum of rewards at each planning step.

\paragraph{Deep generative models.}
Many models in deep learning are used to approximate probability distributions over high dimensional spaces such as spaces of images and videos. 
However, sampling from these distributions requires deep generative models. 
While some deep generative models, such as Generative Adversarial Networks (GANs), try to sample from distributions by making the samples seem as ``realistic" as possible, other models such as Variational Autoencoders (VAEs) map complex distributions to simpler ones via latent space embedding. 
Conditional generative models condition on another variable to sample from conditional distributions~\citep{mirza2014cgan,blurry_vaes}. 
In our MCTS planner, we use deep generative models $G$ to sample observations conditioned on the state, and $P$ to propose state particles given the current observation.  
For our neural network training procedures, we make the ground truth state information available to the planner during training, but keep it unavailable during testing.


\section{Related Works}\label{sec:related}
\paragraph{Planning under uncertainty.}
Planning under state and transition uncertainty requires planners to simultaneously localize and optimally plan through active information gathering.
Popular control-based methods involve variants of the iterative Linear Quadratic Gaussian~\citep{todorov2005ilqg,van2012motion,lee17gpilqg}, which perform trajectory optimization over a simplified belief space.
While such methods can handle continuous dynamics and leverage fast optimization computational tools, they cannot effectively handle image observations and high-dimensional state spaces. They also make simplifying Gaussian belief assumptions and are only locally optimal within the simplified belief space.
Other sampling-based methods such as Sequential Monte Carlo (SMC)~\citep{piche2018smc,wang2020} ameliorate the above shortcomings by augmenting sampling-based planning with advantage networks, but do so at the cost of being unable to reason about future observations.
Such an approximation effectively assumes that the state uncertainty vanishes at the next step, which is proven to be sometimes suboptimal \citep{kaelbling1998planning}.

On the other hand, state-of-the-art tree search planners have shown success in relatively large or continuous space POMDP planning problems.
Most notably, POMCPOW and PFT-DPW \citep{Sunberg2017}, LABECOP \citep{hoerger2020}, and DESPOT-$\alpha$ \citep{Garg2019} were shown to be effective in solving continuous observation POMDP problems.
They use weighted collections of particles to efficiently represent complex beliefs. 
Provided the particles are weighted appropriately based on the observation likelihood, tree search using these particle beliefs will converge to a globally optimal policy \citep{Lim2020}.
Furthermore, these works led to planners that can solve fully continuous or even hybrid space POMDP problems, using techniques such as continuous bandits \citep{lim2021voronoi} or Bayesian optimization \citep{mern2020}.
These tree search methods require access to generative models and observation density models to effectively plan, which we aim to learn with neural networks to extend the scope of the tree search methods.
In this work, we integrate the PFT-DPW algorithm to handle learning-based model components.

\paragraph{Deep learning for vision POMDP.}
Recently, there has been increased interest in solving POMDPs involving visual observations through deep learning~\citep{guillen2005visaugpomdp,karkus2018navnet,zhang2019solar}.
Model-free vision POMDP solvers such as  QMDP-net \citep{Karkus2017} and Deep Variational Reinforcement Learning \citep{igl2018dvrl} maintain a latent belief vector whose update rule is learned via a neural network. They then learn corresponding value or policy networks in this latent belief state and action space. 
Furthermore, since these methods usually make minimal sets of assumptions, extensions of such techniques can also be seen in embodied artificial intelligence applications~\citep{Karkus2021slamnet,ai2022deepvisnav}

Model-based vision POMDP works such as Differentiable Particle Filter (DPF) \citep{Jonschkowski-RSS-2018} allow conventional particle filtering techniques to be interfaced with complex visual observations.
This algorithm contains multiple linked neural network components, including a particle proposer, an observation model, and a dynamics model, that are designed to be trained end-to-end.
A recent work extends DPF with entropy regularization and provides convergence guarantees \citep{pmlr-v139-corenflos21a}.
Dual Sequential Monte Carlo (DualSMC) \citep{wang2020} extends the DPF methods further to introduce an adversarial filtering objective and integrate in the SMC planner, making it a fully closed-loop POMDP solver. 
In this work, we extend DPF and DualSMC to interface tree search planners.

\paragraph{Compositional learning.}
In compositional learning, a learning task is broken down into neural network components that each specialize in different tasks.
These components are then integrated to learn complex relations, allowing for less data and training resources to be used overall. 
Compositional learning can be achieved either through provided compositional structure in the form of an algorithmic prior \citep{Andreas_2016_CVPR,arad2018compositional}, or by automatically discovering such structures \citep{rosenbaum2018routing,pmlr-v87-alet18a,NEURIPS2018_310ce61c,meyerson2018beyond}. 
Specifically, algorithmic prior knowledge can be introduced by specifying neural network architectures or model components, which then allows for  intelligent planning with minimal training from limited or specialized data.
Furthermore, this paradigm has enjoyed success in various reinforcement learning settings, particularly multi-task problems \citep{DBLP:conf/icra/DevinGDAL17,ha2018worldmodels,yang2020multitask,DBLP:conf/icml/MittalLGVSLMB20,lim22transporter}. 

\section{Visual Tree Search}\label{sec:vts}
\begin{figure*}[t]
  \centering
  \includegraphics[width=\textwidth]{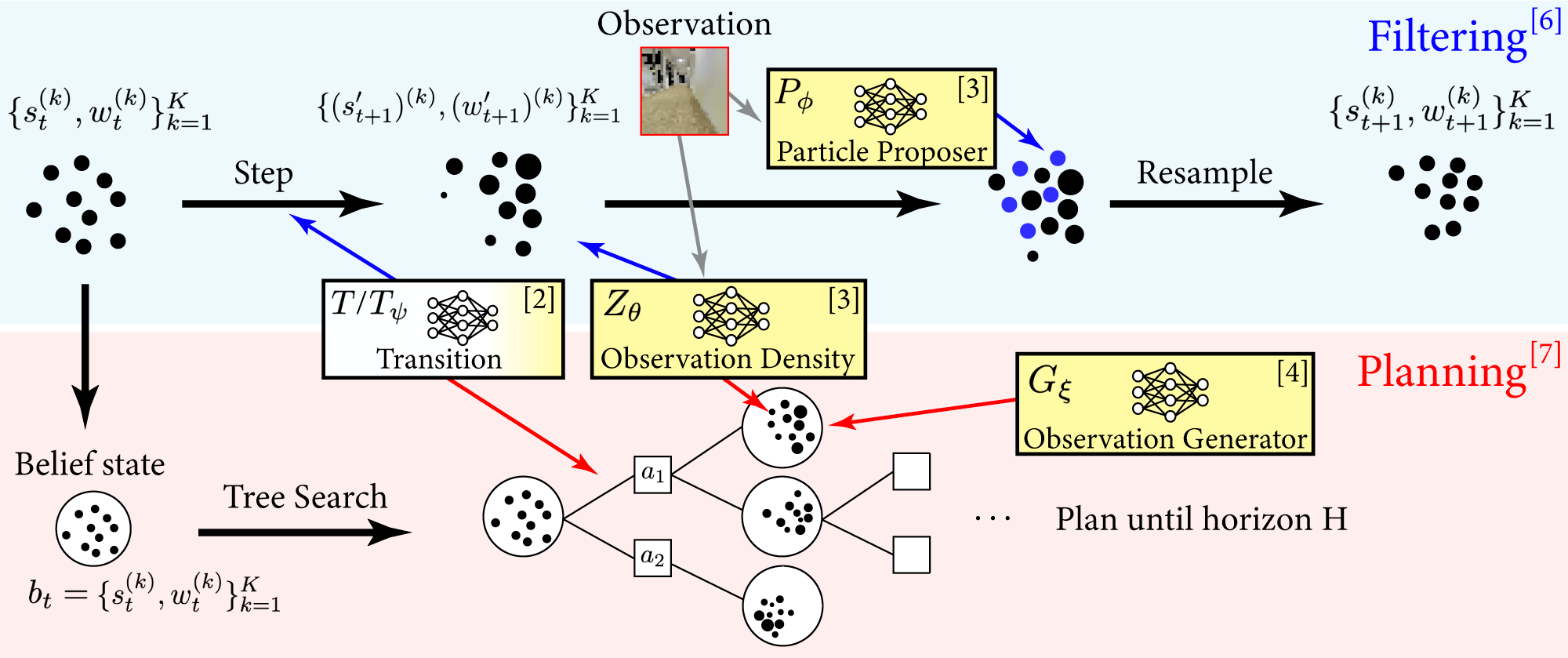}
  \caption{Overview of Visual Tree Search (VTS). 
  The models learned with neural networks are shown in yellow -- the transition model is shown in gradient, since it can be learned or pre-defined. 
  The contributions of the models for filtering are shown in blue arrows, and planning in red arrows, and numbers in brackets show corresponding steps in \cref{alg:vts-algo}.}
  \label{fig:lts}
\end{figure*}
Solving a POMDP can be split into solving the filtering and the planning problems separately. 
In the Visual Tree Search (VTS) algorithm, we integrate the learned POMDP model components with classical filtering and planning techniques.
This results in a POMDP solver that can learn model components that are more sample efficient and interpretable than end-to-end approaches. 
It also benefits from having a robust particle filter and planner built upon techniques with theoretical guarantees, and can adapt to different task rewards.
To integrate planning techniques that use online tree search, we must have access to a conditional generative model that can generate image observations $o$ from a given state $s$ according to the likelihood density $Z(o|s)$.
In this section, we outline the filtering and planning algorithms and models, and the compositional training procedure.

\subsection{Differentiable Particle Filtering}
For particle filtering, we leverage a family of architectures called Differentiable Particle Filters (DPF) \citep{Jonschkowski-RSS-2018}, which combine classical particle filtering algorithms with convolutional neural networks that can handle complex observations.
We learn two neural network-based models: (1) Observation density $Z_\theta$ that gives the likelihood weights $w_t = Z_\theta(o_t|s_t)$, (2) Particle proposer $P_\phi$ that allows us to sample states $s_{t} \sim P_\phi(o_t)$.
The Greek letters denote the parameters of these neural network models. 
Specifically for our work, we adapt the DPF architecture introduced in DualSMC \citep{wang2020}, in which the observation and proposer networks are trained with an adversarial optimization objective.
In principle, we can also train DPF with entropy regularization \citep{pmlr-v139-corenflos21a} with optimality guarantees.
We assume that the transition model $T$ is known, which is not a limiting assumption for many POMDP problems (e.g. POMDPs with physical dynamics), but in principle it can be learned and modeled with a neural network $T_\psi$ as well.

The filtering procedure is described below; the belief is represented by $b_t \approx \{s_t^{(k)}, w_t^{(k)}\}_{k=1}^K$, with $K$ as the number of particles.
First, the agent takes a step with an action chosen by the planner.
Then, the agent updates the predicted states $(s_{t+1}')^{(k)}$ with the transition model $T$.
The observation $o_t$ obtained from the environment is fed into the observation density $Z_\theta$, which provides the likelihood of an observation given the state. 
This is used to update the likelihood weights $(w_{t+1}')^{(k)}$ of each particle.
In order to ensure robustness in the particle representation of the belief state, the particle proposer deep generative model $P_\phi$ proposes plausible state particles for a given observation, replacing some fraction of the particles.
This fraction is made to decay exponentially over time.

\begin{algorithm}[t]
\small
\caption{Visual Tree Search Algorithm.}\label{alg:vts-algo}
\textbf{Input:} Hyperparameters for neural networks ($\zeta$), DPF ($\chi$), MCTS ($\rho$), maximum time step $T_{\max}$.\\
\textbf{Output:} Action $a_t$ at each step $t$.
\begin{algorithmic}[1]
\STATE Collect data $\mathcal{D}$ of tuples $(s_t, a_t, o_t, s_{t+1})$ through random sampling or exploration.
\STATE \textcolor[rgb]{0.6,0.6,0.6}{[Optional]} Train the transition model $T_\psi$ with data tuples $(s_t, a_t, s_{t+1})$.
\STATE \textcolor[rgb]{0.8,0.7,0.0}{[Training]} Jointly train the observation density model $Z_\theta$ and particle proposer model $P_\phi$ with data tuples $(s_t, o_t, \{\hat{s}_{t,i}\})$, where $\{\hat{s}_{t,i}\}$ are particle estimates of $s_t$.
\STATE \textcolor[rgb]{0.8,0.7,0.0}{[Training]} Train the observation conditional generator model $G_\xi$ with data tuples $(s_t, o_t)$.
\FOR{$t=1$ to $T_{\max}$}
    \STATE \textcolor[rgb]{0,0.3,1}{[Filtering]} If $t=1$, initialize the belief state $b_t$. Otherwise, after receiving $o_t$, update the belief state $b_t$ with DPF($T_\psi, Z_\theta, P_\phi, \chi$).
    \STATE \textcolor[rgb]{1,0.1,0.1}{[Planning]} Run MCTS($T_\psi, Z_\theta, G_\xi, \rho$) on the belief state $b_t$ to obtain and perform action $a_t$.
\ENDFOR
\end{algorithmic}
\end{algorithm}
\subsection{Tree Search Planner}
\paragraph{Monte Carlo Tree Search.}\label{sec:mcts}
For the online planner, we use the Particle Filter Trees-Double Progressive Widening (PFT-DPW) algorithm \citep{Sunberg2017}.
PFT-DPW is a particle belief-based MCTS planner that is relatively easy to implement and efficiently vectorizes particle filtering. 
Additionally, a simplified version of PFT-DPW has optimality guarantees \citep{Lim2020}.
However, any continuous POMDP tree search planner can be used instead.

For our navigation problems, we discretize the action space such that the robotic agent moves in the 8 cardinal and diagonal directions with full thrust.
While this means we work with a limited action space, we can ensure that we travel with full thrust to get to the goal faster and reduce the complexity of both planning and generating observations.
However, we could also work with a continuous action space in principle.
Furthermore, we provide PFT-DPW with a na\"ive rollout policy of actuating straight towards the goal and calculating the expected reward, which PFT-DPW can use as a reference and vastly improve upon.

\paragraph{Observation conditional generative model.}\label{sec:generative}
Deep conditional generative models enable online model-based POMDP planning with images.
To plan with a tree search planner, we need to be able to generate the next step states and observations, and evaluate the likelihood of the observations.
With models from DPF, we can generate the next step state with transition model $T$ and calculate the likelihood weight with observation density $Z_\theta$, which are also used in the filtering procedure.
Thus, we only additionally need to learn a deep generative model $G_\xi$ that generates an observation $o_t$ given a state $s_t$: $o_t \sim G_\xi(s_t)$.
We use a Conditional Variational Autoencoder (CVAE) \citep{sohn2015} to model the observation conditional generator $G_\xi$, where the state $s_t$ is the conditional variable, since it had the most consistent training and performance in our experiments. 

\subsection{Compositional Training of Visual Tree Search}
We train each neural network model with pre-collected data $\mathcal{D}$, containing tuples of $(s_t, a_t, o_t, s_{t+1})$, as shown in Lines 1-4 of \cref{alg:vts-algo}.
We provide the  $Z_\theta$ and $P_\phi$ models with randomly sampled batches of state, observation, and synthetic belief particle sets $(s_t, o_t, \{\hat{s}_{t,i}\})$.
While the belief particle sets are not a part of the pre-collected data, we can easily create them by sampling states from a normal distribution centered at $s_t$.
The $Z_\theta$ and $P_\phi$ are trained with the adversarial objective in DualSMC \citep{wang2020}: $Z_\theta$ serves as a discriminator that gives higher likelihoods to states that are more likely for a given observation, and $P_\phi$ serves as a conditional generator that proposes plausible state particles for a given observation.
The $G_\xi$ model is trained with random batches of samples of state and observation pairs: $(s_t, o_t)$.
These random batches provide a good data prior for all types of states and observations an agent may encounter during planning, unlike methods such as DualSMC which only encounter data collected from the locally optimal policy.

Compared to other on-policy RL algorithms and architectures, the VTS training procedure shifts the POMDP problem from a reinforcement learning problem to a compositional learning problem, in which training each model in the POMDP is framed as an unsupervised learning problem.
This drastically decreases the problem complexity, as we have explicit control over the learning objective of each model and the schedule of the training. 
It also allows us to better approximate the data distribution via sampling or exploration, as opposed to an evolving on-policy planner distribution that often starts off poorly and provides heavily biased data.

\section{Experiments}\label{sec:experiments}
\begin{figure*}[t]
  \centering
  \includegraphics[width=\textwidth]{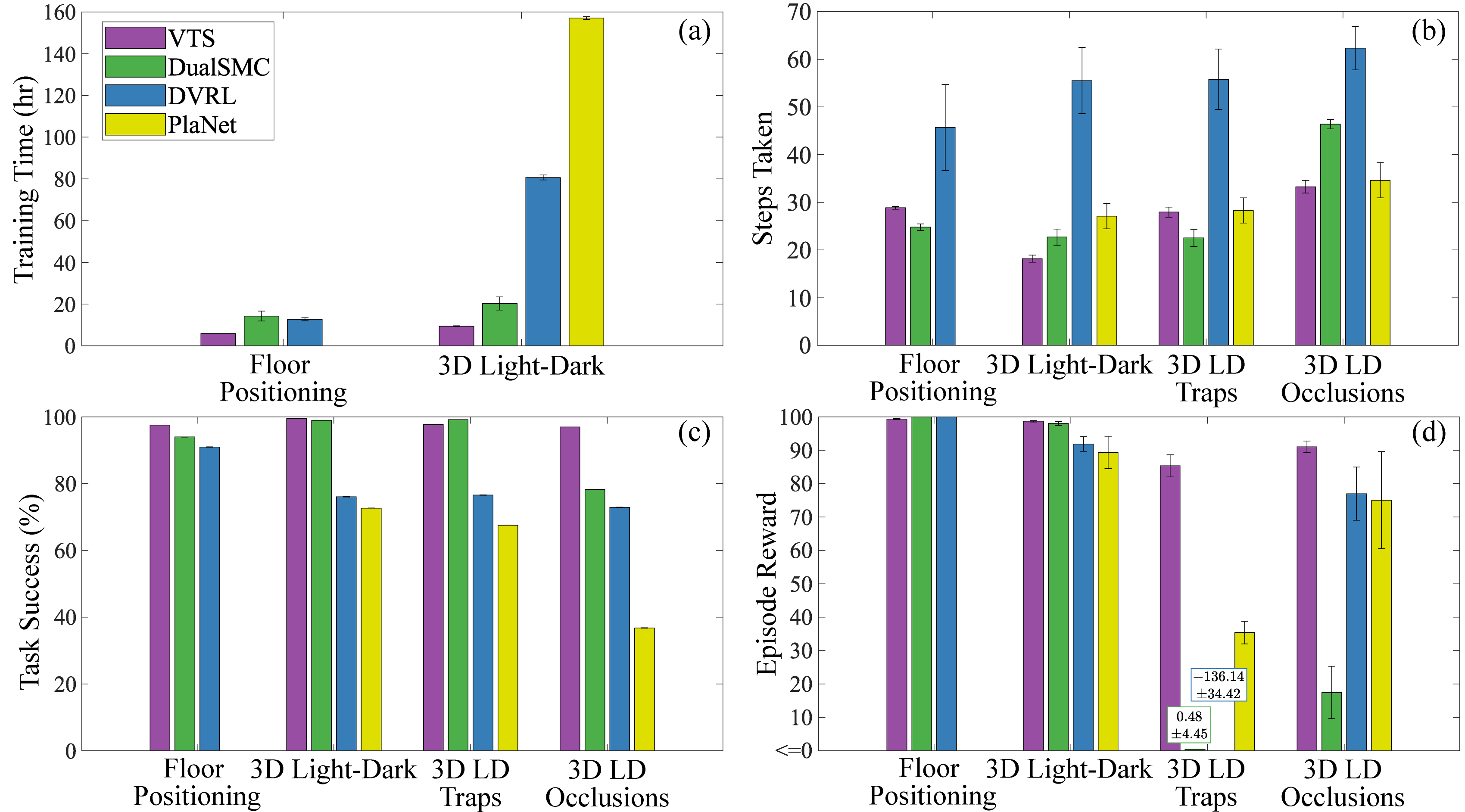}
  \caption{The results of 4 different planners, VTS, DualSMC, DVRL, and PlaNet, on Floor Positioning and 3D Light-Dark. 
  Sub figures denote: (a) training time, (b) steps taken, (c) task success, and (d) episode reward.}
  \label{fig:results}
\end{figure*}

We compare VTS to other state-of-the-art vision POMDP algorithms, DualSMC~\citep{wang2020}, DVRL~\citep{igl2018dvrl} and PlaNet~\citep{hafner2019planet}, on two benchmark vision POMDP problems.
First, we tested our algorithm on the 2D Floor Positioning problem~\citep{wang2020} to demonstrate that using the VTS learning and planning procedure can significantly decrease the training time to learn a solver that is agnostic to the task or reward structure of the problem.
Then, we prepared our own version of the 3D Light-Dark problem in the Stanford Large-Scale 3D Indoor Spaces dataset \citep{armeni2017} to set up a more challenging navigation task that requires the agent to plan with realistic indoor building RGB images.
We also performed ablation tests with the 3D Light-Dark experiment in which we varied the reward structure using spurious traps and the observation space using random visual occlusions.
In each section, we calculate the results of the planner performances for 1000 testing episodes for Floor Positioning and 500 for 3D Light-Dark.
For online planning speed, VTS takes around 0.26 seconds to plan for Floor Positioning and 0.80 seconds for 3D Light-Dark on average, while other planners require less than 0.05 seconds for both problems.
The experimental summary figures are given in \cref{fig:results}. 
In addition, the full tabular summary is given in \cref{app:tabular-summary}, VTS training data details in \cref{app:vts-training}, and the hyperparameters and computation details in \cref{app:hyper}.

\subsection{Floor Positioning Problem}
In this problem, a robotic agent is randomly placed around the center of either the top or the bottom floor, and it must infer its position by relying on a radar-like observation in all four cardinal directions, which bounces off the nearest wall.
The top and bottom floors are indistinguishable within the ``corridor states'' of the hallways, but the robotic agent can take advantage of the ``wall states'' by traveling closer to the top or bottom walls of each floor, where it can receive different observations due to the different wall placements in each floor.
The agent must reach the goal and avoid the trap, where the goal is the left end of the hallway in the top floor and right end in the bottom floor, and the trap is at the opposite side of the goal in each floor.
\begin{figure*}[t]
  \centering
  \subfigure[Initial Belief]{%
    \label{fig:vtsfloor-initial}
    \includegraphics[width=0.32\textwidth]{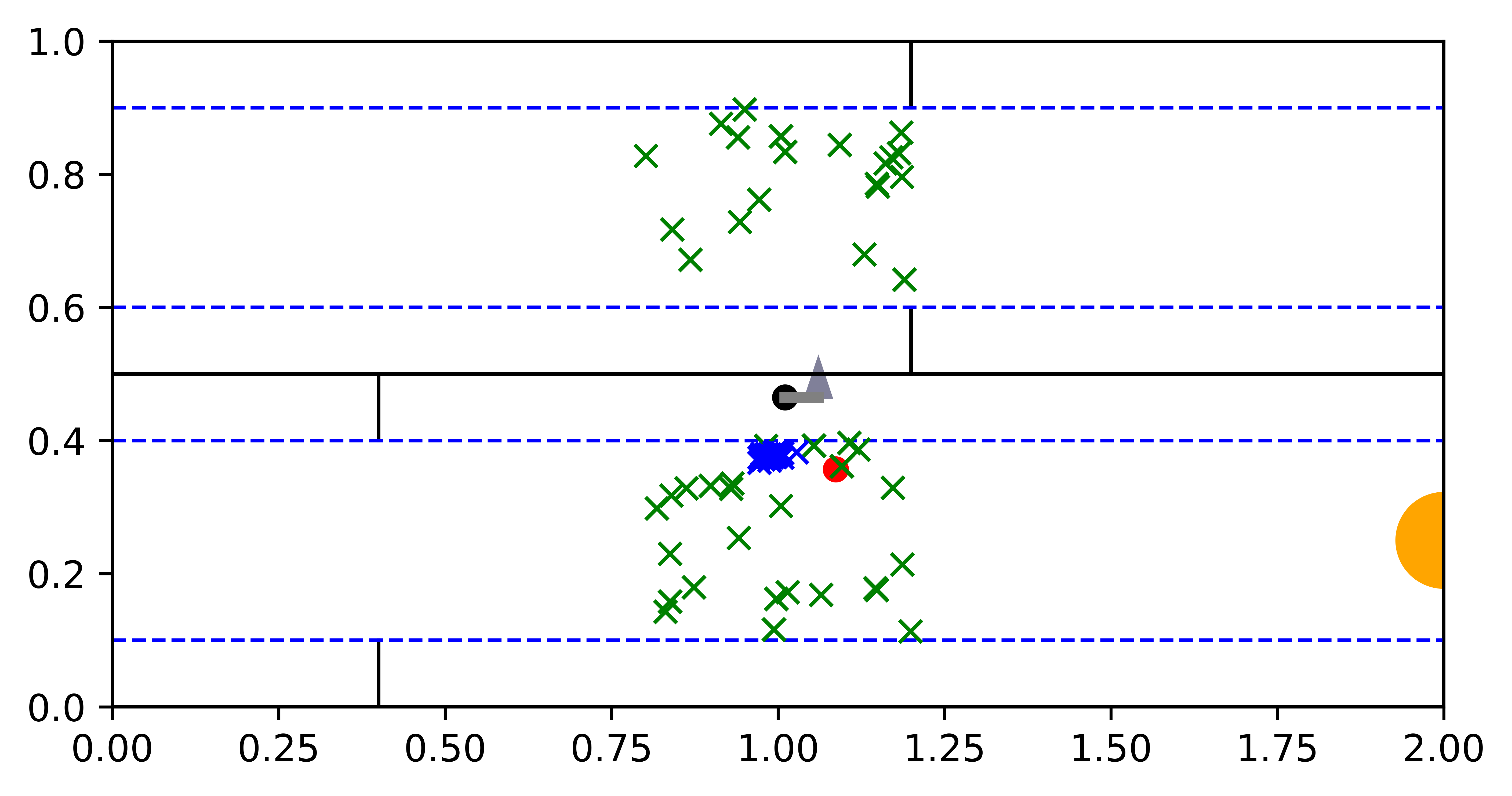}
    }
  \subfigure[Localization]{%
    \label{fig:vtsfloor-localization}
    \includegraphics[width=0.32\textwidth]{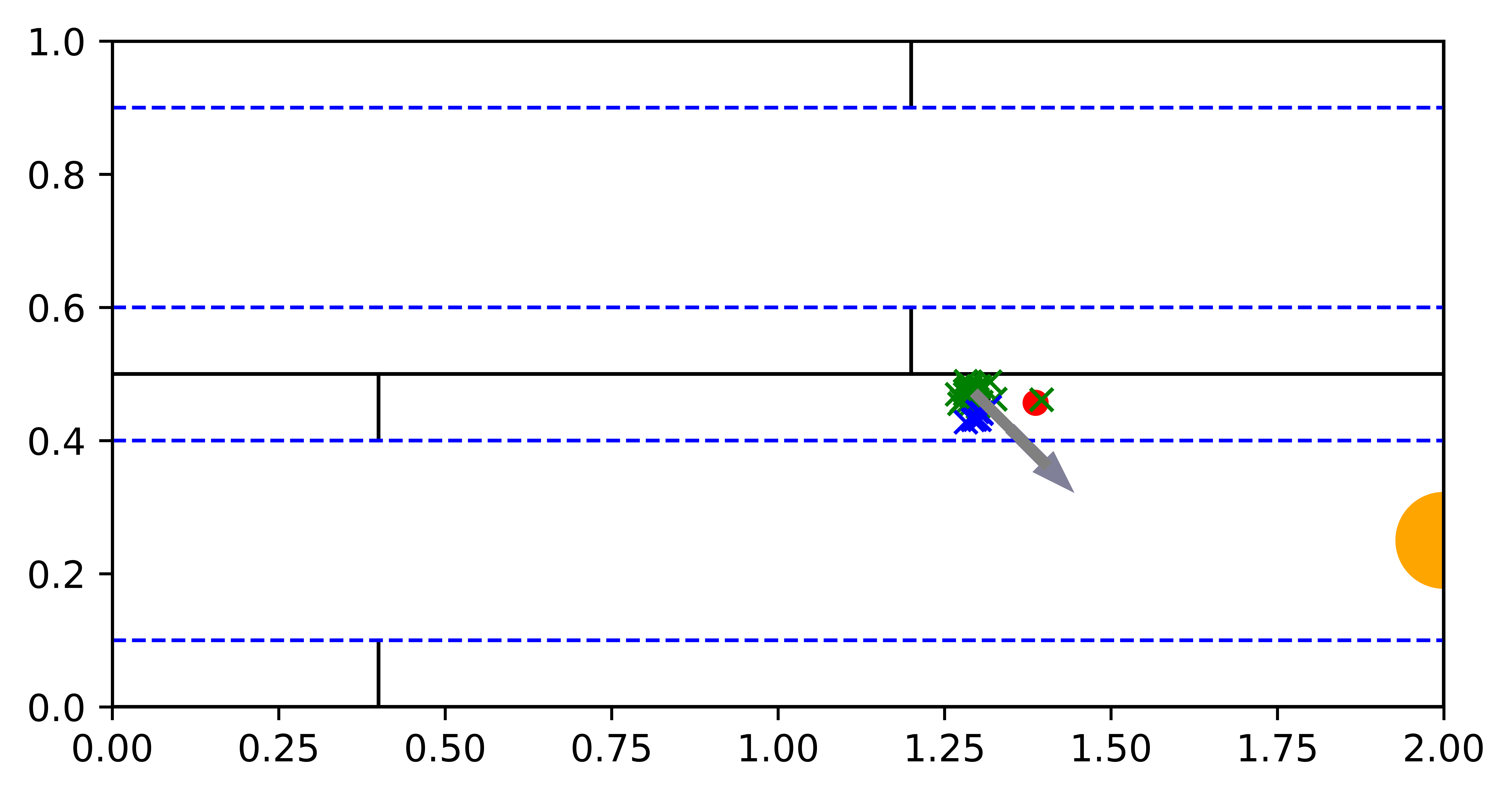}
    }
  \subfigure[Reaching Goal]{%
    \label{fig:vtsfloor-goal}
    \includegraphics[width=0.32\textwidth]{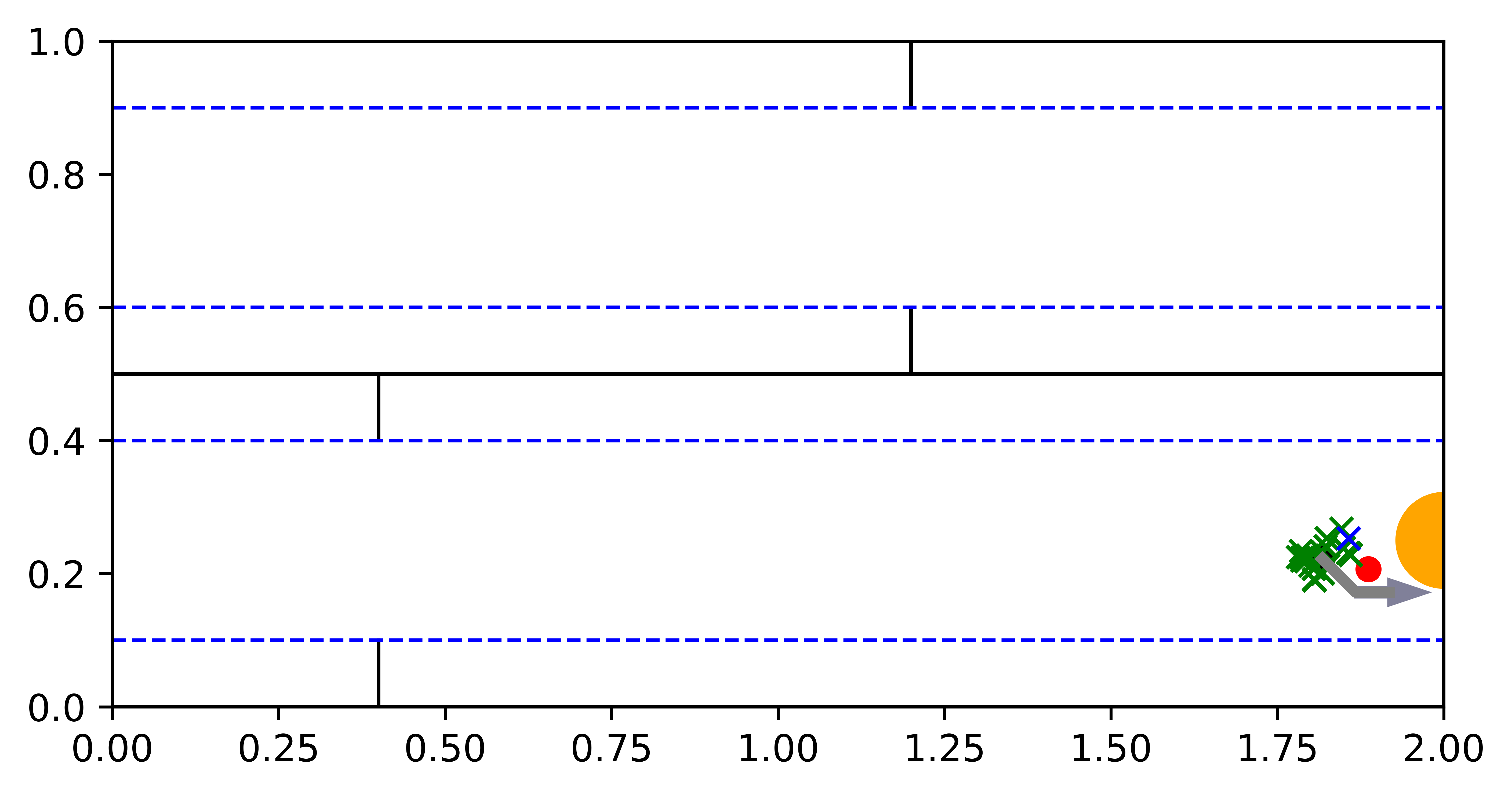}
    }
  \caption{Example VTS behavior for Floor Positioning, where the planner starts the problem with the initial belief (a), localizes in the wall states (b), and successfully reaches the goal (c).}
  \label{fig:vtsfloortraj}
\end{figure*}

\paragraph{Planner comparison.} 
Overall, VTS is the most successful among the three planners with reasonable online planning time and steps taken, while requiring less than half the training time compared to the other planners.
In \cref{fig:vtsfloortraj}, we show an example trajectory of the VTS agent, in which the agent quickly localizes in the wall states and then reaches the goal.

The VTS system gains significant performance optimality and efficiency by balancing offline training and online planning.
Specifically, VTS training is quick since it does not rely on planner performance and only needs to be supplied with relevant state and observation batches.
However, other online training algorithms require the policy to interact with the environment, and as such, much of the time is spent earlier in the training episodes when the planner can neither localize nor reach the goal.
Since tree search methods learn the policy online and are more difficult to parallelize, they typically trade off the offline training time with the online learning and planning time.
Despite the longer planning, VTS plans reasonably fast while maintaining the efficiency and flexibility of an online planner.
We also tested DualSMC with a known $T$ model to ensure VTS is not given an advantage by knowing $T$ for model-based planning, but observed no statistically significant difference in performance.

\subsection{3D Light-Dark Problem}
\begin{wrapfigure}[9]{r}{0.45\textwidth}
    \centering
    \vspace{-0.4cm}
    \includegraphics[width=\linewidth]{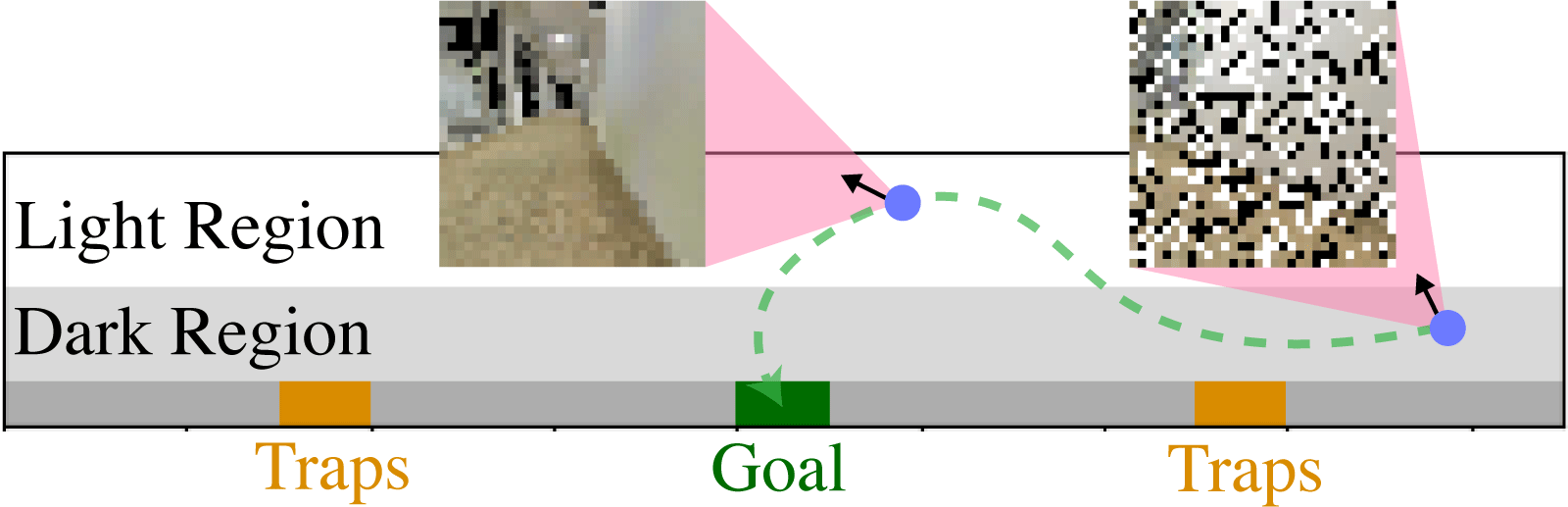}
    \vspace{-0.5cm}
    \caption{Example schema for 3D Light-Dark, localizing in the light region to reach the goal in the dark region.}
  \label{fig:stanford_domain}
\end{wrapfigure}
The Light-Dark problem is a family of problems in which an agent starting in a ``dark'' region can localize by taking a detour into a ``light'' region before reaching the goal. 
Observations are noisier in the dark region than in the light region.
The 3D Light-Dark problem extends this problem to have 3D image observations with salt-and-pepper pixel-wise noise.

The observations consist of $32\times32\times3$ RGB images from the agent's perspective.
Our environment is more challenging to reason with than the one in \cite{wang2020}, since the RGB images are rendered from a realistic hallway scene dataset rather than a synthetic environment.
The agent receives a reward for reaching the goal and a penalty if it enters a ``trap".

\paragraph{Planner comparison.} 
\begin{wrapfigure}[11]{r}{0.45\textwidth}
    \centering
    \vspace{-0.4cm}
    \includegraphics[width=\linewidth]{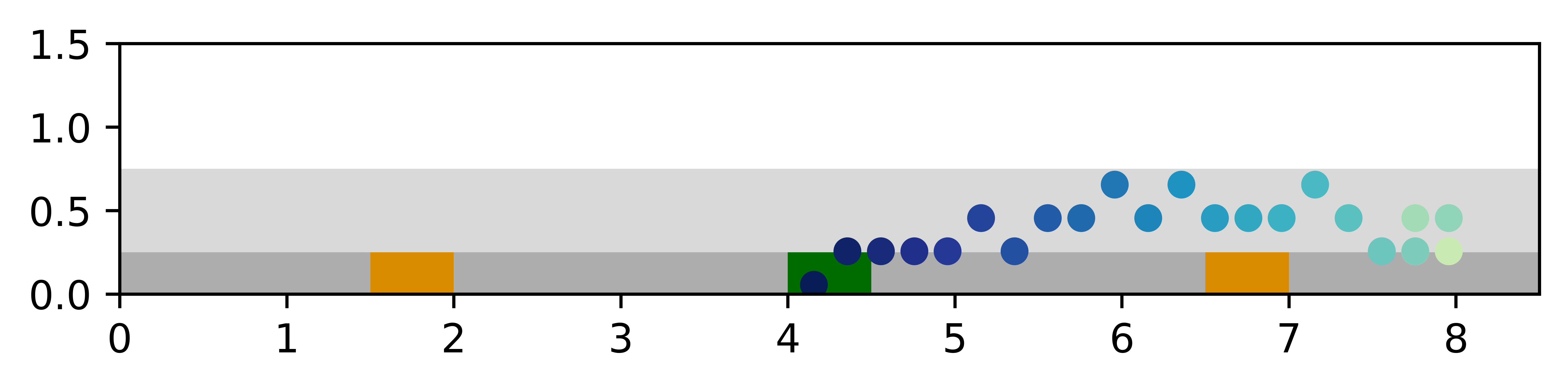}
    \includegraphics[width=\linewidth]{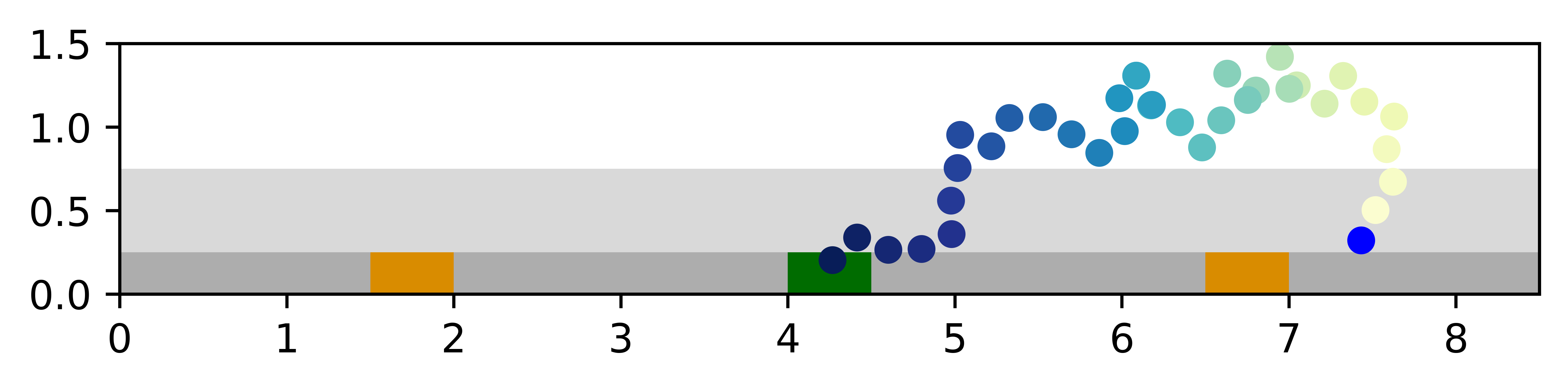}
    \vspace{-0.5cm}
    \caption{VTS can localize without light observations (top), unlike DualSMC (bottom).}
  \label{fig:3dld-vanilla}
\end{wrapfigure}
In the Light-Dark problem and its variants, VTS performs the best among four planners not only by taking full advantage of the image features, but also by being able to adapt to different reward structures and distribution shifts in the noisy observations during test time.
Also, DualSMC sometimes fails to have successful training seeds, and DVRL and PlaNet have lower task success rates despite all successful training seeds.

Among the successful training seeds for DualSMC, VTS and DualSMC have comparable success rates, but the VTS planner takes fewer steps to reach the goal.
While this is only 4.5 steps difference on average, further inspection of the planner trajectories in \cref{fig:3dld-vanilla} reveals an interesting insight.
Unlike the traditional light-dark problems, the VTS results for 3D Light-Dark suggest that the planner in fact is able to localize solely with the noisy observations in the dark region.
This shows that while the salt-and-pepper noise observations are hard to interpret, the corrupted RGB images actually contain sufficient information to localize given enough observations.

\paragraph{Test-time changes: Spurious traps.} 
VTS has additional advantages in its robustness and adaptability, which we showcase through experiments on modified Light-Dark environments.
First, to test the adaptability of different planners on test-time reward structure difference, we perform an ablation with spurious traps that appear during test time.
These regions could represent the presence of unforeseen obstacles or hazards. 
Over 500 testing episodes, we randomly generate the locations of two $0.5 \times 0.5$ square traps over a particular strip in the environment and compare the performances of the models without additional training or modification. 

\begin{wrapfigure}[11]{r}{0.45\textwidth}
    \centering
    \includegraphics[width=\linewidth]{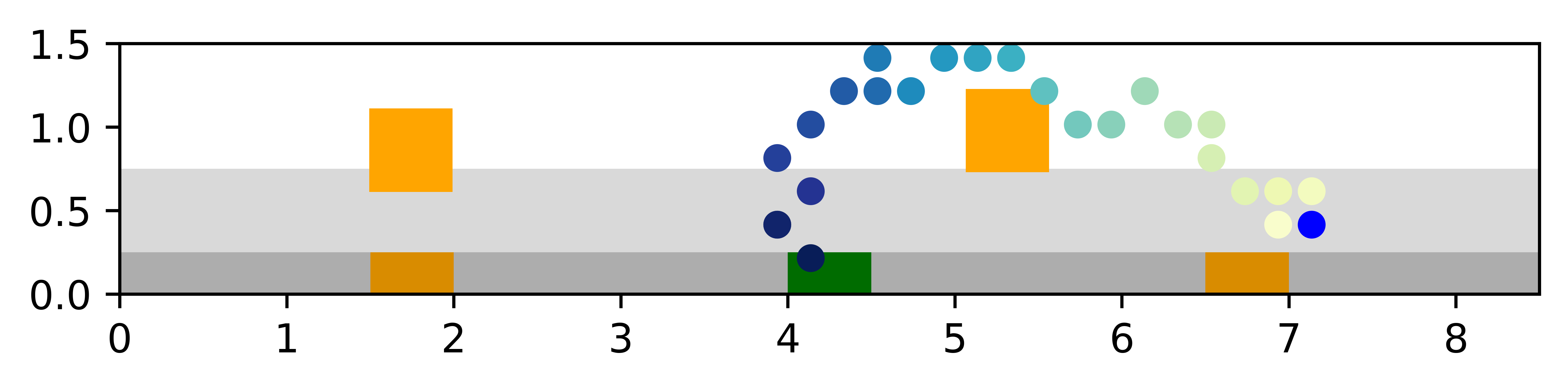}
    \includegraphics[width=\linewidth]{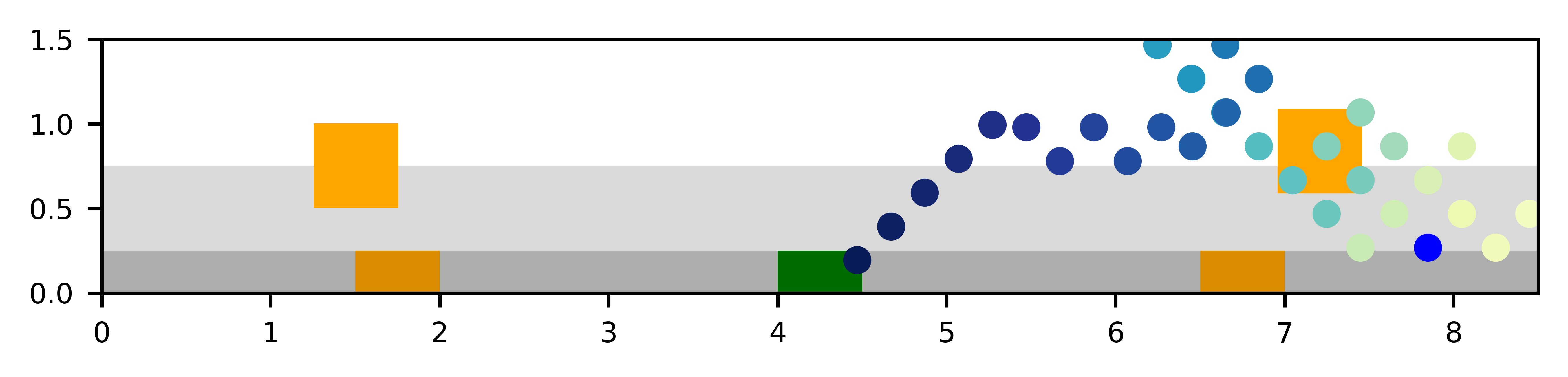}
    \vspace{-0.5cm}
    \caption{VTS can avoid new test time traps (top), while other planners like DVRL disregard the traps (bottom).}
  \label{fig:3dld-traps}
\end{wrapfigure}
Since VTS uses an online planner, it can easily adapt to new reward structures at test time without the need to retrain an entire planner.
We see in \cref{fig:3dld-traps} that while the other planners ignore these new trap regions because they are not represented by their models, the VTS planner is able to take into account rewards seen while planning. 
Thus, while the success rate of all planners remains similar to the vanilla experiment, VTS achieves a much higher reward than all other planners while taking more steps as it actively tries to avoid these traps.

\paragraph{Test-time changes: Image occlusions.} 
Second, we compare planner performances when the form of the noise in the image observations changes during test time. 
During test time only, images seen in the dark region contain random blacked out $15\times15$ squares instead of salt-and-pepper noise. 

\begin{wrapfigure}[12]{r}{0.45\textwidth}
    \centering
    \vspace{-0.4cm}
    \includegraphics[width=\linewidth]{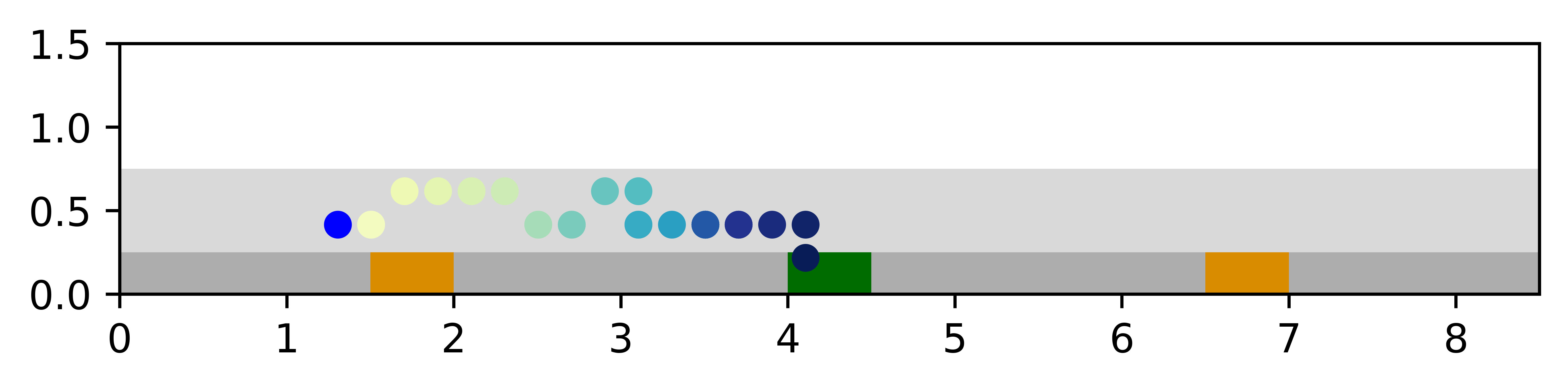}
    \includegraphics[width=\linewidth]{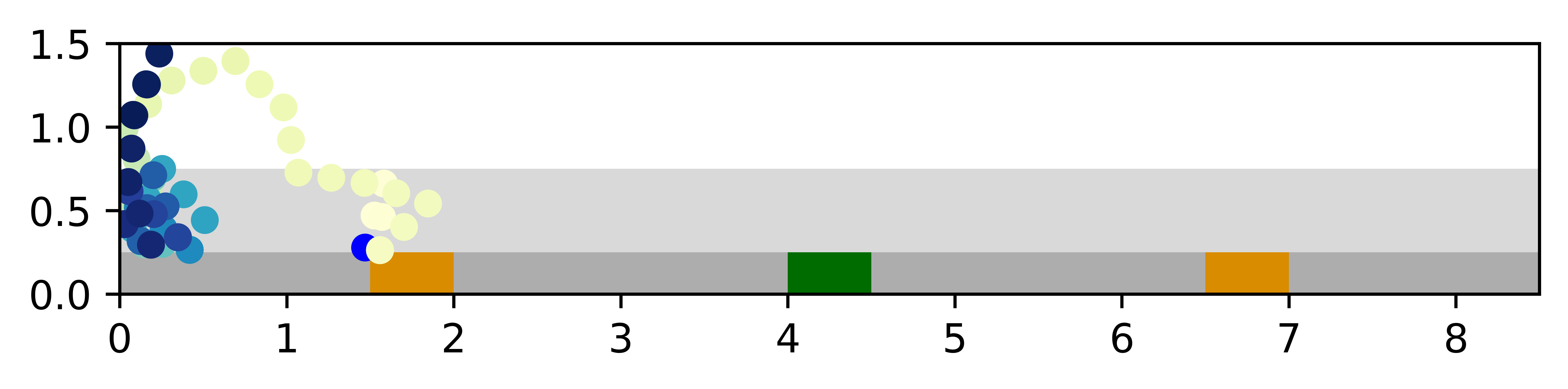}
    \vspace{-0.5cm}
    \caption{VTS can optimally plan even with occlusions (top), while other planners like PlaNet fail to reach the goal (bottom).}
  \label{fig:3dld-occlusion}
\end{wrapfigure}
We also find that VTS is robust to distribution shifts in the image noise at test time.
\cref{fig:3dld-occlusion} shows VTS maintaining a good success rate and number of steps taken, while other planners are struggling to generalize to this new scenario. 
The additional robustness of VTS seems to be due to the $Z_{\theta}$ and $P_{\phi}$ models being trained with high fidelity data samples through random batch sampling or exploration.

In contrast, other planners that use on-policy training and complex non-separable architectures suffer from such distribution shifts.
This is likely due to lack of control over on-policy exploration and the inability to cover many possible scenarios with such limited interaction with the environment.
Due to this difference in data variation, we conjecture that the $Z_{\theta}$ and $P_{\phi}$ models in VTS learn more precise and robust features.


\section{Conclusion}\label{sec:conclusion}
The development of Visual Tree Search suggests new ways to think about integrating uncertainty-aware learning and control.
VTS demonstrates that a more principled integration of control and planning techniques both illuminates the interpretability of each model component and saves training time by cleanly partitioning what each network is responsible for.
This benefits researchers and practitioners alike, as a more interpretable and theoretically principled planner that is also quicker to train is beneficial in many practical safety critical scenarios with limited training resources.

Our work also raises the question of what is the most natural, effective, and safety-ensured method of combining pre-existing controllers and planners, which are extensively studied both theoretically and experimentally, with learning-based components, which can model many complex functions and phenomena.
VTS is one such way of accomplishing that goal, but there are other ways to achieve principled integration of learning and control.

\clearpage
\section*{Acknowledgments}
This material is based upon work supported by a DARPA Assured Autonomy Grant No. FA8750-18-C-0101, the SRC CONIX program No. 2018-JU-2779, NSF CPS Frontiers No. CPS-1545126, the ONR Multibody Control Systems Basic Research Challenge No. N000141812214, Google-BAIR Commons, University of Colorado Boulder, the NASA University Leadership Initiative Grant No. 80NSSC20M0163, and the National Science Foundation Graduate Research Fellowship Program under Grant Nos. DGE 1752814 and DGE 2146752.
Any opinions, findings, and conclusions or recommendations expressed in this material are those of the authors and do not necessarily reflect the views of any aforementioned organizations. 

\bibliography{project.bib}

\begin{thebibliography}{52}
\providecommand{\natexlab}[1]{#1}
\providecommand{\url}[1]{\texttt{#1}}
\expandafter\ifx\csname urlstyle\endcsname\relax
  \providecommand{\doi}[1]{doi: #1}\else
  \providecommand{\doi}{doi: \begingroup \urlstyle{rm}\Url}\fi

\bibitem[Ai et~al.(2022)Ai, Gao, Vinay, and Hsu]{ai2022deepvisnav}
Bo~Ai, Wei Gao, Vinay, and David Hsu.
\newblock Deep visual navigation under partial observability.
\newblock In \emph{2022 International Conference on Robotics and Automation
  (ICRA)}, pages 9439--9446, 2022.

\bibitem[Alet et~al.(2018)Alet, Lozano-Perez, and Kaelbling]{pmlr-v87-alet18a}
Ferran Alet, Tomas Lozano-Perez, and Leslie~P. Kaelbling.
\newblock Modular meta-learning.
\newblock volume~87 of \emph{PMLR}, pages 856--868. PMLR, 29--31 Oct 2018.

\bibitem[Andreas et~al.(2016)Andreas, Rohrbach, Darrell, and
  Klein]{Andreas_2016_CVPR}
Jacob Andreas, Marcus Rohrbach, Trevor Darrell, and Dan Klein.
\newblock Neural module networks.
\newblock In \emph{IEEE CVPR}, June 2016.

\bibitem[{Armeni} et~al.(2017){Armeni}, {Sax}, {Zamir}, and
  {Savarese}]{armeni2017}
I.~{Armeni}, A.~{Sax}, A.~R. {Zamir}, and S.~{Savarese}.
\newblock {Joint 2D-3D-Semantic Data for Indoor Scene Understanding}.
\newblock \emph{ArXiv e-prints}, February 2017.

\bibitem[Ayer et~al.(2012)Ayer, Alagoz, and Stout]{ayer2012mammography}
Turgay Ayer, Oguzhan Alagoz, and Natasha~K. Stout.
\newblock A {POMDP} approach to personalize mammography screening decisions.
\newblock \emph{Operations Research}, 60\penalty0 (5):\penalty0 1019--1034,
  2012.

\bibitem[Bai et~al.(2015)Bai, Cai, Ye, Hsu, and Lee]{bai2015intention}
Haoyu Bai, Shaojun Cai, Nan Ye, David Hsu, and Wee~Sun Lee.
\newblock Intention-aware online {POMDP} planning for autonomous driving in a
  crowd.
\newblock In \emph{IEEE ICRA}, pages 454--460, Seattle, WA, USA, 2015. {IEEE}.

\bibitem[{B}ertsekas(2005)]{bertsekas2005dynamic}
{D}. {B}ertsekas.
\newblock \emph{{D}ynamic {P}rogramming and {O}ptimal {C}ontrol}.
\newblock Massachusetts: Athena Scientific, 2005.

\bibitem[{Browne} et~al.(2012){Browne}, {Powley}, {Whitehouse}, {Lucas},
  {Cowling}, {Rohlfshagen}, {Tavener}, {Perez}, {Samothrakis}, and
  {Colton}]{browne2012}
C.~B. {Browne}, E.~{Powley}, D.~{Whitehouse}, S.~M. {Lucas}, P.~I. {Cowling},
  P.~{Rohlfshagen}, S.~{Tavener}, D.~{Perez}, S.~{Samothrakis}, and
  S.~{Colton}.
\newblock A survey of monte carlo tree search methods.
\newblock \emph{IEEE T-CIAIG}, 4\penalty0 (1):\penalty0 1--43, 2012.

\bibitem[Corenflos et~al.(2021)Corenflos, Thornton, Deligiannidis, and
  Doucet]{pmlr-v139-corenflos21a}
Adrien Corenflos, James Thornton, George Deligiannidis, and Arnaud Doucet.
\newblock Differentiable particle filtering via entropy-regularized optimal
  transport.
\newblock In \emph{ICML}, volume 139 of \emph{PMLR}, pages 2100--2111. PMLR,
  18--24 Jul 2021.

\bibitem[Devin et~al.(2017)Devin, Gupta, Darrell, Abbeel, and
  Levine]{DBLP:conf/icra/DevinGDAL17}
Coline Devin, Abhishek Gupta, Trevor Darrell, Pieter Abbeel, and Sergey Levine.
\newblock Learning modular neural network policies for multi-task and
  multi-robot transfer.
\newblock In \emph{IEEE ICRA}, pages 2169--2176. {IEEE}, 2017.

\bibitem[Garg et~al.(2019)Garg, Hsu, and Lee]{Garg2019}
Neha~P. Garg, David Hsu, and Wee~Sun Lee.
\newblock {DESPOT}-$\alpha$: Online {POMDP} planning with large state and
  observation spaces.
\newblock In \emph{RSS}, 2019.

\bibitem[Guill{\'{e}}n et~al.(2005)Guill{\'{e}}n, Bergasa, Barea, and
  Escudero]{guillen2005visaugpomdp}
Mar{\'{\i}}a Elena~L{\'{o}}pez Guill{\'{e}}n, Luis~Miguel Bergasa, Rafael
  Barea, and Mar{\'{\i}}a~Soledad Escudero.
\newblock A navigation system for assistant robots using visually augmented
  pomdps.
\newblock \emph{Auton. Robots}, 19\penalty0 (1):\penalty0 67--87, 2005.

\bibitem[Ha and Schmidhuber(2018)]{ha2018worldmodels}
David Ha and J{\"u}rgen Schmidhuber.
\newblock Recurrent world models facilitate policy evolution.
\newblock In \emph{NeuRIPS}, pages 2451--2463. Curran Associates, Inc., 2018.

\bibitem[Hafner et~al.(2019)Hafner, Lillicrap, Fischer, Villegas, Ha, Lee, and
  Davidson]{hafner2019planet}
Danijar Hafner, Timothy Lillicrap, Ian Fischer, Ruben Villegas, David Ha,
  Honglak Lee, and James Davidson.
\newblock Learning latent dynamics for planning from pixels.
\newblock In \emph{International Conference on Machine Learning}, pages
  2555--2565, 2019.

\bibitem[Hoerger and Kurniawati(2020)]{hoerger2020}
Marcus Hoerger and Hanna Kurniawati.
\newblock An on-line {POMDP} solver for continuous observation spaces.
\newblock \emph{ArXiv e-prints}, 2020.

\bibitem[Holland et~al.(2013)Holland, Kochenderfer, and
  Olson]{holand2013optimizing}
Jessica~E. Holland, Mykel~J. Kochenderfer, and Wesley~A. Olson.
\newblock Optimizing the next generation collision avoidance system for safe,
  suitable, and acceptable operational performance.
\newblock \emph{{A}ir {T}raffic {C}ontrol {Q}uarterly}, 21\penalty0
  (3):\penalty0 275--297, 2013.

\bibitem[Hudson and Manning(2018)]{arad2018compositional}
Drew~Arad Hudson and Christopher~D. Manning.
\newblock Compositional attention networks for machine reasoning.
\newblock In \emph{ICLR}, 2018.

\bibitem[Igl et~al.(2018)Igl, Zintgraf, Le, Wood, and Whiteson]{igl2018dvrl}
Maximilian Igl, Luisa Zintgraf, Tuan~Anh Le, Frank Wood, and Shimon Whiteson.
\newblock Deep variational reinforcement learning for {POMDP}s.
\newblock In \emph{ICML}, volume~80 of \emph{PMLR}, pages 2117--2126. PMLR,
  10--15 Jul 2018.

\bibitem[Jonschkowski et~al.(2018)Jonschkowski, Rastogi, and
  Brock]{Jonschkowski-RSS-2018}
Rico Jonschkowski, Divyam Rastogi, and Oliver Brock.
\newblock Differentiable particle filters: End-to-end learning with algorithmic
  priors.
\newblock In \emph{RSS}, Pittsburgh, Pennsylvania, June 2018.

\bibitem[Kaelbling et~al.(1998)Kaelbling, Littman, and
  Cassandra]{kaelbling1998planning}
Leslie~Pack Kaelbling, Michael~L. Littman, and Anthony~R. Cassandra.
\newblock Planning and acting in partially observable stochastic domains.
\newblock \emph{Artificial Intelligence}, 101\penalty0 (1):\penalty0 99 -- 134,
  1998.

\bibitem[Karkus et~al.(2017)Karkus, Hsu, and Lee]{Karkus2017}
Peter Karkus, David Hsu, and Wee~Sun Lee.
\newblock Qmdp-net: Deep learning for planning under partial observability.
\newblock In \emph{NeuRIPS}, volume~30. Curran Associates, Inc., 2017.

\bibitem[Karkus et~al.(2018)Karkus, Hsu, and Lee]{karkus2018navnet}
Peter Karkus, David Hsu, and Wee~Sun Lee.
\newblock Integrating algorithmic planning and deep learning for partially
  observable navigation.
\newblock \emph{ArXiv e-prints}, 2018.

\bibitem[Karkus et~al.(2020)Karkus, Angelova, Vanhoucke, and
  Jonschkowski]{karkus2020dmnet}
Peter Karkus, Anelia Angelova, Vincent Vanhoucke, and Rico Jonschkowski.
\newblock Differentiable mapping networks: Learning structured map
  representations for sparse visual localization.
\newblock In \emph{IEEE ICRA}, pages 4753--4759, 2020.

\bibitem[Karkus et~al.(2021)Karkus, Cai, and Hsu]{Karkus2021slamnet}
Peter Karkus, Shaojun Cai, and David Hsu.
\newblock Differentiable slam-net: Learning particle slam for visual
  navigation.
\newblock In \emph{IEEE CVPR}, pages 2815--2825, June 2021.

\bibitem[Kirsch et~al.(2018)Kirsch, Kunze, and Barber]{NEURIPS2018_310ce61c}
Louis Kirsch, Julius Kunze, and David Barber.
\newblock Modular networks: Learning to decompose neural computation.
\newblock In \emph{NeuRIPS}, volume~31. Curran Associates, Inc., 2018.

\bibitem[Kochenderfer(2015)]{kochenderfer2015decision}
Mykel~J. Kochenderfer.
\newblock \emph{Decision Making Under Uncertainty: Theory and Application}.
\newblock Massachusetts: {MIT} Press, 2015.

\bibitem[Kurniawati and Yadav(2016)]{kurniawati2016online}
Hanna Kurniawati and Vinay Yadav.
\newblock An online {POMDP} solver for uncertainty planning in dynamic
  environment.
\newblock In \emph{Robotics Research}, pages 611--629. Springer, 2016.

\bibitem[Lee et~al.(2017)Lee, Srinivasa, and Mason]{lee17gpilqg}
Gilwoo Lee, Siddhartha~S. Srinivasa, and Matthew~T. Mason.
\newblock {GP-ILQG}: Data-driven robust optimal control for uncertain nonlinear
  dynamical systems.
\newblock \emph{ArXiv e-prints}, 2017.

\bibitem[Lim et~al.(2020)Lim, Tomlin, and Sunberg]{Lim2020}
Michael~H. Lim, Claire Tomlin, and Zachary~N. Sunberg.
\newblock Sparse tree search optimality guarantees in {POMDP}s with continuous
  observation spaces.
\newblock In \emph{IJCAI}, pages 4135--4142. International Joint Conferences on
  Artificial Intelligence, Inc., 7 2020.

\bibitem[Lim et~al.(2021)Lim, Tomlin, and Sunberg]{lim2021voronoi}
Michael~H. Lim, Claire~J. Tomlin, and Zachary~N. Sunberg.
\newblock Voronoi progressive widening: Efficient online solvers for continuous
  state, action, and observation {POMDP}s.
\newblock In \emph{IEEE CDC}, pages 4493--4500, 2021.

\bibitem[Lim et~al.(2022)Lim, Zeng, Ichter, Bandari, Coumans, Tomlin, Schaal,
  and Faust]{lim22transporter}
Michael~H. Lim, Andy Zeng, Brian Ichter, Maryam Bandari, Erwin Coumans, Claire
  Tomlin, Stefan Schaal, and Aleksandra Faust.
\newblock Multi-task learning with sequence-conditioned transporter networks.
\newblock In \emph{IEEE ICRA}, pages 2489--2496, 2022.

\bibitem[Mern et~al.(2021)Mern, Yildiz, Sunberg, Mukerji, and
  Kochenderfer]{mern2020}
John Mern, Anil Yildiz, Zachary Sunberg, Tapan Mukerji, and Mykel~J.
  Kochenderfer.
\newblock Bayesian optimized {M}onte {C}arlo planning.
\newblock In \emph{AAAI}. AAAI Press, 2021.

\bibitem[Meyerson and Miikkulainen(2018)]{meyerson2018beyond}
Elliot Meyerson and Risto Miikkulainen.
\newblock Beyond shared hierarchies: Deep multitask learning through soft layer
  ordering.
\newblock In \emph{ICLR}, 2018.

\bibitem[Mirza and Osindero(2014)]{mirza2014cgan}
Mehdi Mirza and Simon Osindero.
\newblock Conditional generative adversarial nets.
\newblock \emph{ArXiv e-prints}, 2014.

\bibitem[Mittal et~al.(2020)Mittal, Lamb, Goyal, Voleti, Shanahan, Lajoie,
  Mozer, and Bengio]{DBLP:conf/icml/MittalLGVSLMB20}
Sarthak Mittal, Alex Lamb, Anirudh Goyal, Vikram Voleti, Murray Shanahan,
  Guillaume Lajoie, Michael Mozer, and Yoshua Bengio.
\newblock Learning to combine top-down and bottom-up signals in recurrent
  neural networks with attention over modules.
\newblock In \emph{ICML}, volume 119 of \emph{PMLR}, pages 6972--6986. {PMLR},
  2020.

\bibitem[Mnih et~al.(2013)Mnih, Kavukcuoglu, Silver, Graves, Antonoglou,
  Wierstra, and Riedmiller]{mnih-atari-2013}
Volodymyr Mnih, Koray Kavukcuoglu, David Silver, Alex Graves, Ioannis
  Antonoglou, Daan Wierstra, and Martin Riedmiller.
\newblock Playing atari with deep reinforcement learning.
\newblock In \emph{NeurIPS Deep Learning Workshop}. 2013.

\bibitem[Papadimitriou and Tsitsiklis(1987)]{papadimitriou1987complexity}
Christos~H. Papadimitriou and John~N. Tsitsiklis.
\newblock The complexity of {M}arkov decision processes.
\newblock \emph{Mathematics of Operations Research}, 12\penalty0 (3):\penalty0
  441--450, 1987.

\bibitem[Pich\'{e} et~al.(2019)Pich\'{e}, Thomas, Ibrahim, Bengio, and
  Pal]{piche2018smc}
Alexandre Pich\'{e}, Valentin Thomas, Cyril Ibrahim, Yoshua Bengio, and Chris
  Pal.
\newblock Probabilistic planning with sequential monte carlo methods.
\newblock In \emph{ICLR}, 2019.

\bibitem[Rosenbaum et~al.(2018)Rosenbaum, Klinger, and
  Riemer]{rosenbaum2018routing}
Clemens Rosenbaum, Tim Klinger, and Matthew Riemer.
\newblock Routing networks: Adaptive selection of non-linear functions for
  multi-task learning.
\newblock In \emph{ICLR}, 2018.

\bibitem[Silver and Veness(2010)]{Silver2010}
David Silver and Joel Veness.
\newblock {M}onte-{C}arlo planning in large {POMDP}s.
\newblock In \emph{NeuRIPS}, pages 2164--2172. Curran Associates, Inc., 2010.

\bibitem[Singh et~al.(2021)Singh, Peri, Kim, Kim, and Ahn]{singh21worldbelief}
Gautam Singh, Skand Peri, Junghyun Kim, Hyunseok Kim, and Sungjin Ahn.
\newblock Structured world belief for reinforcement learning in {POMDP}.
\newblock In \emph{ICML}, volume 139, pages 9744--9755. PMLR, 18--24 Jul 2021.

\bibitem[Sohn et~al.(2015)Sohn, Lee, and Yan]{sohn2015}
Kihyuk Sohn, Honglak Lee, and Xinchen Yan.
\newblock Learning structured output representation using deep conditional
  generative models.
\newblock In \emph{NeuRIPS}, volume~28. Curran Associates, Inc., 2015.

\bibitem[Sunberg and Kochenderfer(2018)]{Sunberg2017}
Zachary Sunberg and Mykel~J. Kochenderfer.
\newblock Online algorithms for {POMDP}s with continuous state, action, and
  observation spaces.
\newblock In \emph{ICAPS}, Delft, Netherlands, 2018. AAAI Press.

\bibitem[Sunberg et~al.(2017)Sunberg, Ho, and Kochenderfer]{sunberg2017value}
Zachary~N. Sunberg, Christopher~J. Ho, and Mykel~J. Kochenderfer.
\newblock The value of inferring the internal state of traffic participants for
  autonomous freeway driving.
\newblock In \emph{ACC}, pages 3004--3010, Seattle, WA, USA, 2017. IEEE.

\bibitem[Todorov and Li(2005)]{todorov2005ilqg}
E.~Todorov and Weiwei Li.
\newblock A generalized iterative lqg method for locally-optimal feedback
  control of constrained nonlinear stochastic systems.
\newblock In \emph{ACC}, volume~1, pages 300--306, 2005.

\bibitem[Van Den~Berg et~al.(2012)Van Den~Berg, Patil, and
  Alterovitz]{van2012motion}
Jur Van Den~Berg, Sachin Patil, and Ron Alterovitz.
\newblock Motion planning under uncertainty using iterative local optimization
  in belief space.
\newblock \emph{International Journal of Robotics Research}, 31\penalty0
  (11):\penalty0 1263--1278, 2012.

\bibitem[Wang et~al.(2020)Wang, Liu, Wu, Zhu, Du, Fei-Fei, and
  Tenenbaum]{wang2020}
Yunbo Wang, Bo~Liu, Jiajun Wu, Yuke Zhu, Simon~S. Du, Li~Fei-Fei, and Joshua~B.
  Tenenbaum.
\newblock Dualsmc: Tunneling differentiable filtering and planning under
  continuous pomdps.
\newblock In \emph{IJCAI}, pages 4190--4198. International Joint Conferences on
  Artificial Intelligence Organization, 7 2020.

\bibitem[Yang et~al.(2020)Yang, Xu, Wu, and Wang]{yang2020multitask}
Ruihan Yang, Huazhe Xu, Yi~Wu, and Xiaolong Wang.
\newblock Multi-task reinforcement learning with soft modularization.
\newblock \emph{ArXiv e-prints}, 2020.

\bibitem[Ye et~al.(2017)Ye, Somani, Hsu, and Lee]{ye2017despot}
Nan Ye, Adhiraj Somani, David Hsu, and Wee~Sun Lee.
\newblock {DESPOT}: Online {POMDP} planning with regularization.
\newblock \emph{Journal of Artificial Intelligence Research}, 58:\penalty0
  231--266, 2017.

\bibitem[Young et~al.(2013)Young, Ga{\v{s}}i{\'c}, Thomson, and
  Williams]{young2013pomdp}
Steve Young, Milica Ga{\v{s}}i{\'c}, Blaise Thomson, and Jason~D Williams.
\newblock {POMDP}-based statistical spoken dialog systems: A review.
\newblock \emph{IEEE}, 101\penalty0 (5):\penalty0 1160--1179, 2013.

\bibitem[Zhang et~al.(2019)Zhang, Vikram, Smith, Abbeel, Johnson, and
  Levine]{zhang2019solar}
Marvin Zhang, Sharad Vikram, Laura~M. Smith, Pieter Abbeel, Matthew~J. Johnson,
  and Sergey Levine.
\newblock {SOLAR:} deep structured representations for model-based
  reinforcement learning.
\newblock In \emph{ICML}, volume~97, pages 7444--7453, 2019.

\bibitem[Zhao et~al.(2017)Zhao, Song, and Ermon]{blurry_vaes}
Shengjia Zhao, Jiaming Song, and Stefano Ermon.
\newblock Towards deeper understanding of variational autoencoding models.
\newblock \emph{ArXiv e-prints}, 2017.

\end{thebibliography}

\clearpage
\onecolumn
\appendix
\section*{Appendix}

\section{Tabular Summary of Experimental Results}\label{app:tabular-summary}
\begin{table*}[h]
  \centering
  \footnotesize
  \caption{Summary of planner performances on the Floor Positioning problem and the 3D Light-Dark problem variants. Each column reports the mean of means estimator over different seeds, and $\pm$ indicates one standard error of the estimator. Mean particle distance is defined as the mean of the distance between the mean belief state and the true state. All statistics are for when the agent is successful.}
  \label{tab:results}
  \begin{tabular}{lralalal}
    \toprule
    \multicolumn{2}{l}{\textbf{Planner \& Problem}} & \textbf{Task} & \textbf{Episode} & \textbf{Steps} & \textbf{Training} & \textbf{Planning} & \textbf{Particle} \\
    \multicolumn{2}{l}{(Successful Trials)} & \textbf{Success (\%)} & \textbf{Reward} & \textbf{Taken} & \textbf{Time (h)} & \textbf{Time (s)} & \textbf{Distance}\\
    \midrule
    \multicolumn{2}{l}{\textcolor{gray}{\textbf{Floor Positioning}}} &&&&&& \\
    \textbf{VTS} & \textcolor{gray}{(10/10)} & \textbf{97.6} \tiny{$\pm 0.002$} & 99.36 \tiny{$\pm$ 0.12} & 28.85 \tiny{$\pm 0.28$} & \textbf{5.88} \tiny{$\pm 0.02$} & 0.26 \tiny{$\pm 0.0022$}  & \textbf{0.07} \tiny{$\pm 0.002$}  \\
    DualSMC & \textcolor{gray}{(10/10)} & 94.0 \tiny{$\pm 0.049$} & \textbf{100.0} \tiny{$\pm 0.0$} & \textbf{24.78} \tiny{$\pm 0.70$} & 14.27 \tiny{$\pm 2.39$} & 0.04 \tiny{$\pm 0.0003$} & 0.08 \tiny{$\pm 0.003$} \\
    DVRL & \textcolor{gray}{(10/10)} & 91.0 \tiny{$\pm 0.084$} & \textbf{100.0} \tiny{$\pm 0.0$} & 45.71 \tiny{$\pm 9.01$} & 12.74 \tiny{$\pm 0.72$} & 0.0063 \tiny{$\pm 6.4$e-5} & -- \\
    \midrule
    \multicolumn{2}{l}{\textcolor{gray}{\textbf{3D Light-Dark}}} &&&&&& \\
    \textbf{VTS} & \textcolor{gray}{(10/10)} & \textbf{99.6} \tiny{$\pm 0.002$} & \textbf{98.71} \tiny{$\pm$ 0.22} & \textbf{18.18} \tiny{$\pm$ 0.76} & \textbf{9.42} \tiny{$\pm 0.21$} & 0.76 \tiny{$\pm$ 0.005} & \textbf{0.48} \tiny{$\pm$ 0.07} \\
    DualSMC & \textcolor{gray}{(\textbf{6}/10)} & 99.0 \tiny{$\pm 0.009$} & 98.05 \tiny{$\pm$ 0.62} & 22.71 \tiny{$\pm$ 1.70} & 20.31 \tiny{$\pm 3.13$} & 0.06 \tiny{$\pm$ 0.003} & 0.50 \tiny{$\pm$ 0.02} \\
    DVRL & \textcolor{gray}{(10/10)} & 76.1 \tiny{$\pm 0.081$} & 91.88 \tiny{$\pm$ 2.2} & 55.33 \tiny{$\pm$ 6.94} & 80.65 \tiny{$\pm 1.18$} & 0.0076 \tiny{$\pm 3.0$e-4} & -- \\
    PlaNet & \textcolor{gray}{(4/4)} & 72.7 \tiny{$\pm 0.029$} & 89.35 \tiny{$\pm$ 4.84} & 27.1 \tiny{$\pm$ 2.67} & 157.1 \tiny{$\pm 0.66$} & 0.061 \tiny{$\pm 4.1$e-4} & -- \\
    \midrule
    \multicolumn{2}{l}{\textcolor{gray}{\textbf{3D LD + Traps}}}&&&&&&\\
    \textbf{VTS} & & 97.7 \tiny{$\pm 0.003$} & \textbf{85.34} \tiny{$\pm 3.33$} & 27.96 \tiny{$\pm 1.06$} & -- & 0.80 \tiny{$\pm 0.0039$} & \textbf{0.42} \tiny{$\pm 0.06$}  \\
    DualSMC & & \textbf{99.2} \tiny{$\pm 0.007$} & 0.48 \tiny{$\pm 4.45$} & \textbf{22.55} \tiny{$\pm 1.80$} & -- & 0.06 \tiny{$\pm 0.0006$} & 0.49 \tiny{$\pm 0.02$} \\
    DVRL & & 76.6 \tiny{$\pm 0.078$} & -136.14 \tiny{$\pm 34.42$} & 55.79 \tiny{$\pm 6.34$} & -- & 0.0076 \tiny{$\pm 2.9$e-4} & -- \\
    PlaNet & & 67.6 \tiny{$\pm 0.016$} & 35.41 \tiny{$\pm 3.38$} & 28.31 \tiny{$\pm 2.63$} & -- & 0.061 \tiny{$\pm 7.1$e-4} & -- \\
    \midrule
    \multicolumn{2}{l}{\textcolor{gray}{\textbf{3D LD + Occlusions}}}&&&&&&\\
    \textbf{VTS} & & \textbf{97.0} \tiny{$\pm 0.007$} & \textbf{91.05} \tiny{$\pm 1.73$} & \textbf{33.25} \tiny{$\pm 1.33$} & -- & 0.77 \tiny{$\pm 0.0051$} & \textbf{0.82} \tiny{$\pm 0.08$} \\
    DualSMC & & 78.3 \tiny{$\pm 0.052$} & 17.45 \tiny{$\pm 7.82$} & 46.38 \tiny{$\pm 0.97$} & -- & 0.06 \tiny{$\pm 0.0005$} & 0.91 \tiny{$\pm 0.03$} \\
    DVRL & & 72.9 \tiny{$\pm 0.081$} & 77.0 \tiny{$\pm 7.97$} & 62.32 \tiny{$\pm 4.56$} & -- & 0.0075 \tiny{$\pm 3.1$e-4} & -- \\
    PlaNet & & 36.8 \tiny{$\pm 0.078$} & 75.07 \tiny{$\pm 14.55$} & 34.62 \tiny{$\pm 3.69$} & -- & 0.059 \tiny{$\pm 5.4$e-4} & -- \\
    \bottomrule
  \end{tabular}
\end{table*}

For all methods, the episode length was $200$ steps. 
If the agent reached the goal in under $200$ steps, this was considered a success with a positive reward of $100$ and a termination of the episode. 
Entrance into traps incurred a negative reward of $-100$ per entry.
This is with the exception of DVRL, in which rewards were scaled to lie between $-1$ and $1$ to ensure stability of the training. 
The DVRL rewards reported here are scaled by $100$.

In addition to statistics such as the success rate, reward, and so on, we have provided values for the number of successful trials for the method. 
This is because certain seeds for DualSMC failed to produce any goal-reaching behavior at all during training. 
Since the provided statistics are only for episodes in which the goal was reached, those seeds of DualSMC were dropped entirely.

The prohibitive training time of PlaNet allowed for only $4$ trials to be run with reasonable computational resources. 
Training for all methods was done on NVIDIA Tesla K80 GPUs.

Note that the ``planning time" reported for DVRL is actually the time to run the forward pass of the DVRL policy, since the method does not perform planning. 
The number has been provided to display that aspect of the algorithm.
\clearpage

\section{VTS Training Data Generation Details}\label{app:vts-training}
The open source code for our implementations will be available at \url{https://github.com/michaelhlim/VisualTreeSearch}.

\subsection{Floor Positioning}
We trained the $Z_\theta, P_\phi$, and $G_\xi$ models with random data batches.
$Z_\theta$ and  $P_\phi$ were provided with random batches of samples $(s_t, o_t, \{\hat{s}_{t,i}\})$, with $s_t = (x_t, y_t)$ being the agent's position. The $\{\hat{s}_{t,i}\}$ were generated from a normal distribution centered at $s_t$ with standard deviation 0.01. 
$G_\xi$ was provided with random batches of samples $(s_t, o_t)$, where for each batch we sampled the $s_t$ and generated each $o_t$ from the corresponding $s_t$.
The sampling procedure for the states was biased towards the wall states, since those states were more difficult to learn from. 
Specifically, for $Z_\theta$ and  $P_\phi$, in each batch a state would be chosen from a wall with $0.5$ probability and from anywhere in the environment with $0.5$ probability.
For $G_\xi$, in each batch a state would be chosen from a wall with $0.5$ probability, from a non-wall area with $0.375$ probability, and from anywhere in the environment with $0.125$ probability.

\subsection{3D Light-Dark}
To train the $Z_{\theta}$ and $P_{\phi}$ models, we employed the following procedure.
Each training batch consisted of random batches of samples $(s_t, \theta_t, o_t, \{\hat{s}_{t, i}\})$, where $s_t = (x_t, y_t)$.
The agent's location could be described by both its position $s_t$ and its heading $\theta_t$, though only $s_t$ was unknown.
In each batch, $\theta_t$ was discretely chosen from $[0, 2\pi)$ at intervals of $\pi/4$.
This is because the agent's actions were in the form of discrete commands instantaneously changing its heading.
If $s_t$ was in the dark region, then the $\{\hat{s}_{t, i}\}$ were generated from a normal distribution centered at $s_t$ with standard deviation 0.1, and 0.01 for the light region.
The difference in standard deviation was to enable the $Z_{\theta}$ and $P_{\phi}$ models to cause better localization in the light region than in the dark region. 
The two models were trained on a schedule that gradually introduced noise in the image observations $o_t$ over the training epochs. 
In contrast, $G_{\xi}$ was trained simply with the random batches $(s_t, \theta_t, o_t)$ without the noise schedule, as if it were data collected by a randomly exploring agent.
However, both sets of models could be trained either way in principle.
For these conditional generative models, the concatenation of $s_t$ and $\theta_t$ formed the conditional variable.

\subsection{On-Policy Refining}
Technically, we could employ additional on-policy learning to refine the networks to focus on parts of the problem that are relevant for a given task as done in training DPF-based policies \citep{Jonschkowski-RSS-2018}.
However, we found that on-policy learning does not significantly improve performance in addition to regular training, since the planner often successfully and quickly reaches the goal and does not provide many new samples in areas where the models need improvement.

\clearpage
\section{Hyperparameters and Computation Details}\label{app:hyper}

\cref{tab:pretraining} shows the hyperparameters used in training and filtering for VTS.
Our models were trained on NVIDIA Tesla K80 GPUs.
The hyperparameters for DualSMC training remained the same as the original work \citep{wang2020}.
For 3D Light-Dark, we maintained a training pool of data to draw random batches of samples from instead of generating image observations at every training step.

\begin{table}[h]
    \centering
    \caption{Summary of hyperparameters used in training and filtering.}
    \label{tab:pretraining}
    \begin{tabular}{lll}
        \toprule
        Hyperparameters & Floor Positioning & 3D Light-Dark\\
        \midrule
        Training gradient steps & 500,000 & 100,000 \\
        Training sample pool size & - & 16,000\\
        Learning rate & 0.001 & 0.0003 \\
        Batch size & 64 & 64\\
        \midrule
        Filtering particles & 100 & 100 \\
        Percent of particles proposed & 0.3 & 0.3 \\
        Resampling frequency & 3 & 3 \\
        \bottomrule
    \end{tabular}
\end{table}

Unlike DualSMC, VTS does not use an LSTM as one of the intermediate layers in $Z_\theta$ because $Z_\theta$ is trained offline from randomly collected data. 
We observed that this would result in particle de-localization, as the observation density could not propagate forward the localized information. 
In essence, the VTS solver would be quick to localize, but sometimes would need to take a detour in order to re-localize.
Particle de-localization has been observed before in the literature and there are many potential methods to remedy this problem. 
We found an exponential decay of the number of particles proposed, as suggested by \cite{wang2020}, to provide the best results and solve the issue.
That is, for episode step number $n$ and nominal particle proposal amount $p$, the number of particles proposed is $p * \gamma_d^n$.
We set the decay rate $\gamma_d$ to $0.9$.   

\cref{tab:pftdpw} shows the hyperparameters used for PFT-DPW planner.
While hyperparameters for tree search planners are often chosen with hyperparameter sweeps and/or optimization, such meta-optimization in our algorithm would mean having to run our algorithm on a GPU numerous times, which was practically very inefficient.
Thus, the hyperparameters were manually chosen via a combination of inspection and prior experience.
For both Floor Positioning and 3D Light-Dark, we found the rollout strategy that yielded the best results to be one in which we assume that the uncertainty in the current belief collapses in the next step.
Essentially, we select a random particle from the belief and calculate its straight-line direction and distance to the goal.
We use this distance to estimate the discounted reward associated with that particle.
We add that with the weighted sum of rewards obtained when all other particles move in that same direction and distance. 

\cref{tab:environments} shows some parameters for the Floor Positioning and 3D Light-Dark environments. 
The environment parameters for Floor Positioning were left unchanged from \cite{wang2020}.

Across all baseline planning methods, the dimension of the action varied according to the problem in accordance with \cite{wang2020}; for the Floor Positioning problem, the action dimension was $2$, representing a change in the 2D state. 
For the 3D Light-Dark problem, the action dimension was $1$, representing a change in the agent's orientation. 
DVRL, however, was found to produce the best performance when the action was 2D for both problems.

\begin{table}[h]
    \centering
    \caption{Summary of hyperparameters used in PFT-DPW.}
    \label{tab:pftdpw}
    \begin{tabular}{cccccccccc}
        \toprule
        & $n$ & $c$ & $k_a$ & $\alpha_a$ & $k_o$ & $\alpha_o$ & $m$ & $H$ & $\gamma$\\ 
        \midrule
        \textbf{Floor Positioning} & & & & & & \\
        & 100 & 10.0 & 3.0 & $\frac{1}{4}$ & 4.0 & $\frac{1}{4}$ & 100 & 10 & 0.99\\
        \midrule
        \textbf{3D Light-Dark} & & & & & & \\
         & 100 & 10.0 & 3.0 & $\frac{1}{4}$ & 4.0 & $\frac{1}{4}$ & 100 & 10 & 0.99\\
        \bottomrule
    \end{tabular}
\end{table}

\begin{table}[h]
    \centering
    \caption{Summary of environment parameters.}
    \label{tab:environments}
    \begin{tabular}{lll}
        \toprule
        Hyperparameters & Floor Positioning & 3D Light-Dark\\
        \midrule
        Observation size & 4 & 32 $\times$ 32 $\times$ 3 \\
        Agent velocity & 0.05 & 0.2 \\
        Noise amount (image) & - & 40$\%$ \\
        Occlusion amount (image) & - & 15 $\times$ 15 \\
        \bottomrule
    \end{tabular}
\end{table}

For the 3D Light-Dark problem, the $Z_\theta$ and $P_\phi$ modules were trained with a gradual noise schedule that was as follows: for the first $1/4$ of the number of training epochs, the images in the training data had $0 \%$ noise. For the next $1/4$ of the epochs, the images corresponding to states in the dark region had $final \; noise \; \%* 1/4$ noise. For the next $1/4$ of the epochs, the images had $final \; noise \; \% * 1/2$ noise. For the final $1/4$ of the epochs, the images had the full $final \; noise \; \%$. (As seen in Table \ref{tab:environments}, $final \; noise \; \%$ was $40 \%$.) This percentage corresponds to the percent of locations on the $32 \times 32$ image that were corrupted by the salt-and-pepper noise.

Finally, \cref{tab:neuralnet} shows the neural network architecture details for the $Z, P, G$ models we trained on different experimental domains.
We note here that although VTS and DualSMC differ in the architectures of their $Z_{\theta}$ and $P_{\phi}$ models in that the networks for VTS have deeper structures than those for DualSMC, we did not see any noticeable increase in performance for DualSMC with the deeper architectures in a preliminary exploration.

\begin{table}[h]
    \centering
    \caption{Network details.}
    \label{tab:neuralnet}
    \begin{tabular}{lll}
        \toprule
        Model & Layers & \# Channels \\ 
        \midrule
        \textbf{Floor Positioning} & & \\
        $Z_\theta$ & 5-layer MLP & $256\times 3,128, 16$ \\
         & Concat: state & $18$ \\
         & 3-layer MLP & $256\times 2, 1$ \\
        $P_\phi$ & 3-layer MLP & $256\times 2, 64$\\
         & Concat: $z(64) \sim \mathcal{N}(0,1)$ & $128$\\
         & 4-layer MLP & $256\times 3, 2$\\
        $G_\xi$ \; \; \; \; \; Encoder & 5-layer MLP  & $256 \times 4$, 64 \\
        & 1 layer to output $\mu$, $\sigma$ & 64 \\
        \; \; \; \; \; \; \; \; Decoder & 5-layer MLP & $256 \times 4$, 4\\
        
        \midrule
        
        \textbf{3D Light-Dark} & & \\
        $Z_\theta$ & Conv2d, filter (3, 3), stride 1 & num image channels \\
        & Conv2d $\times$ 4, filter (3, 3), stride 2 & 32, 64, 128, 256 \\
        & Conv2d, filter (3, 3), stride 1 & 512 \\
        & 3-layer MLP  & 1024, 512, 256  \\
        & Normalize output (*) & 256 \\
        & 3-layer MLP & 256 $\times$ 3 \\
        & Concat: state, orientation & 259 \\
        & 4-layer MLP & 128 $\times$ 3, 1 \\
        $P_\phi$ & Same as $Z_\theta$ up to (*) & 256 \\
        & 1-layer MLP & 256 \\
        & Concat: $z(256) \sim \mathcal{N}(0,1)$, orientation & 513 \\ 
        & 3-layer MLP & 128 $\times$ 2, 2 \\
        $G_\xi$ \; \; \; \; \; Encoder & Same as $Z_\theta$ up to (*) & 256 \\
        & Concat: state, orientation & 259 \\
        & 5-layer MLP & 256 $\times$ 4, 128 \\
        & 1 layer MLP to output $\mu$, $\sigma$ & 128 \\
        \; \; \; \; \; \; \; \; Decoder & Concat $z(128) \sim \mathcal{N}(0,1)$, state, orientation & 131\\
        & 5-layer MLP & 256 $\times$ 5 \\
        & 3-layer MLP  & 256, 512, 1024  \\
        & Conv2dTranspose $\times$ 3, filter (3, 3), stride 2 & 128, 64, 32 \\
        & Conv2d, filter (3, 3), stride 2 & num image channels \\
        \bottomrule
    \end{tabular}
\end{table}

\clearpage
\section{Stanford 3D Light-Dark Environment Implementation Details}\label{app:stanford3d}
\begin{figure*}[h]
  \centering
  \includegraphics[width=0.45\textwidth]{figures/stanford-environment-diagram-2.png}
  \includegraphics[width=0.50\textwidth]{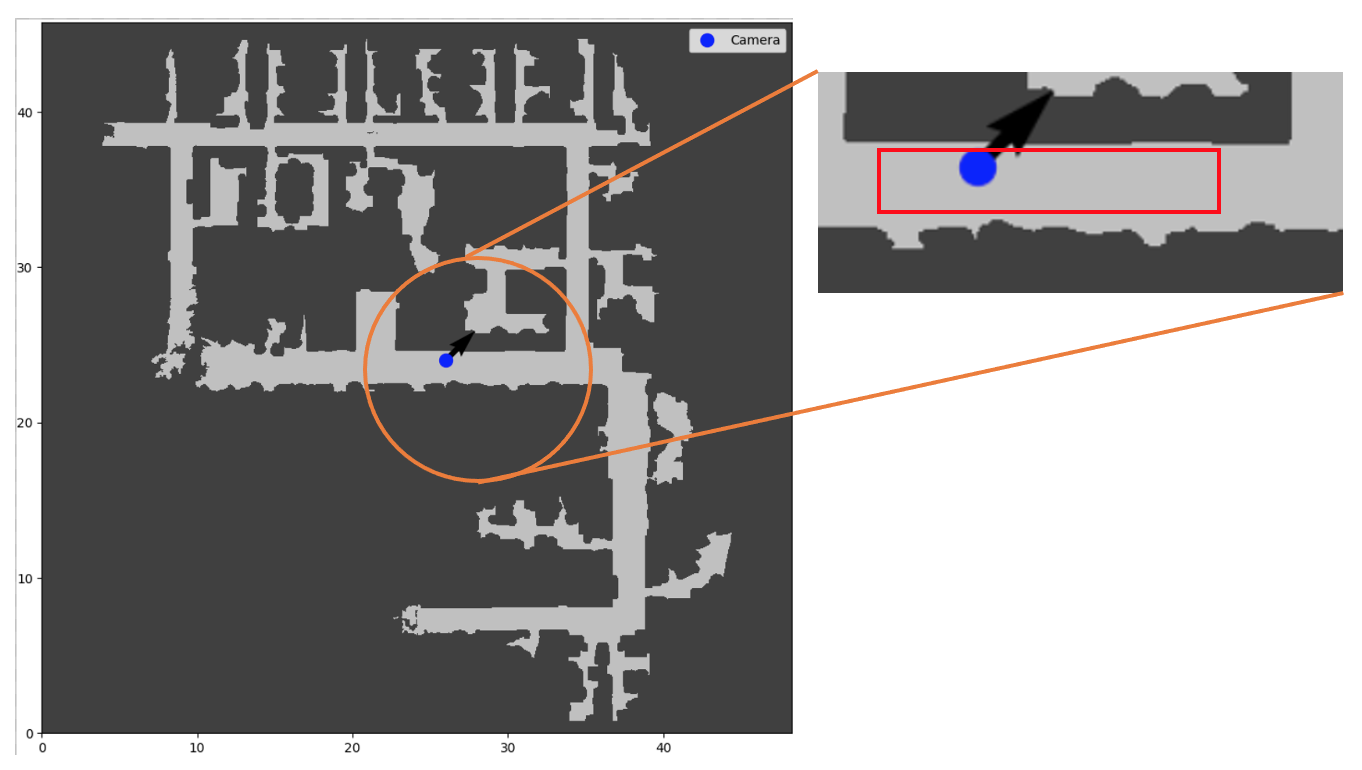}
  \caption{The overview diagram of the 3D Light-Dark problem implemented using the Stanford dataset (left), and a top-down map of a floor in a building in the Stanford dataset that we use to recreate the 3D Light-Dark problem (right).}
  \label{fig:stanford_full}
\end{figure*}

\cref{fig:stanford_full} shows the overview diagram of the 3D Light-Dark problem implemented using the Stanford dataset (left), and a top-down map of a floor in a building in the Stanford dataset (right).

The environment diagram depicts the ``trap" regions in orange, the goal region in green, and the agent in blue. 
The agent's location is given by $(x, y, \theta)$, where $x,y$ are the unknown absolute position coordinates and $\theta$ is the known heading. 
The shaded part of the environment depicts the ``dark" region, where the image observations received by the agent are corrupted by salt-and-pepper noise.
The rest of the environment is in the ``light", where the image observations are uncorrupted.
The darkest gray region is a wall region that the agent cannot enter.
The agent's initial state is randomly chosen from a strip in the dark region.

In the top-down map, the red rectangle denotes the part of the building that forms our environment (top right). 
We chose a subset of the Stanford dataset region in order to replicate the exact settings of the 3D Light-Dark problem in \cite{wang2020}.
However, since our 3D Light-Dark implementation uses realistic images rendered with the Stanford dataset as opposed to using synthetic backgrounds generated with the DeepMind simulation platform, our problem is more realistic and challenging.

\clearpage
\section{Baseline Planner Implementation Details}\label{app:baseline}
We compare the performance of VTS against two baselines in addition to DualSMC: Deep Variational Reinforcement Learning for POMDPs (DVRL), and Deep Planning Network (PlaNet).

We evaluated DVRL in both the Floor Positioning and 3D Light-Dark environments and PlaNet in the 3D Light-Dark environment. 
This was so that we could minimally modify the model architectures provided by the two methods; while DVRL architectures supported both vector and image observations, the PlaNet architecture only supported image observations. 
This was also the reason for running the PlaNet 3D Light-Dark experiments with $64 \times 64$ images while all other baselines and VTS used $32 \times 32$ images. 
In this way, the baselines were advantaged in our comparisons.

\subsection{DVRL}
Table \ref{tab:dvrl} shows the parameters modified when running DVRL, with all other parameters unchanged from the original work. 
For Floor Positioning, a sweep was performed over the encoder channel dimensions, presence of the BatchNorm, multiplier backprop length, and number of particles.
For 3D Light-Dark, a sweep was performed over the number of environment processes, number of environment steps per gradient step, and multiplier backprop length to balance performance and computation time. 

Since the DVRL implementation is designed to produce trajectories in the environment by running multiple environment processes in parallel, we performed both training and testing in this multiprocessing framework. 
For testing, we ran $2$ environment processes in parallel and collected the $1000$ testing episodes for Floor Positioning and $500$ testing episodes for 3D Light-Dark from their combined results. 
Also, the DVRL policy produces actions for these parallel environments at once.
Therefore, the ``planning time" provided is the time for the policy to produce $2$ actions. 

\begin{table}[h]
    \centering
    \caption{Summary of DVRL environment, training, and model hyperparameters.}
    \label{tab:dvrl}
    \begin{tabular}{lll}
        \toprule
        Hyperparameters & Floor Positioning & 3D Light-Dark\\
        \midrule
        \textbf{Environment} &  &  \\
        Observation size & 4 & 32 $\times$ 32 $\times$ 3 \\
        Agent velocity & 0.05 & 0.2 \\
        Noise amount (image) & -- & 40$\%$ \\
        Occlusion amount (image) & -- & 15 $\times$ 15 \\
        \midrule
        \textbf{Training} &  &  \\
        Number of training frames & 2.5e7 & 1.2e6 \\
        Number of env steps per gradient step & 25 & 5 \\
        Multiplier backprop length & 1 & 100 \\
        Number of particles & 10 & 10 \\
        Number of environment processes & 16 & 4 \\
        \midrule
        \textbf{Model} &  &  \\
        Encoder channel dimensions & $[64, 64]$ & $[32, 64, 32]$ \\
        BatchNorm present & Yes & Yes \\
        RNN latent state size ($h$) & 256 &  256\\
        Additional latent state size ($z$) & 256 & 256 \\
        Action encoding size & 64 & 128 \\
        \bottomrule
    \end{tabular}
\end{table}

\subsection{PlaNet}

Table \ref{tab:planet} shows the parameters modified when running PlaNet, with all other parameters unchanged from the original work.
A sweep was performed over action repeat and learning rate.
The occlusion amount was increased to $30$ as compared to $15$ for the other baselines to maintain consistency with the larger image size.

Since the action repeat for the PlaNet baseline is 2, the agent only needs to perform the planning, which is done via the Cross Entropy Method (CEM), every 2 steps. 
This is because the action resulting from the plan is applied twice. 
To reflect this aspect of the algorithm in the results, we divided the obtained planning time by 2 to produce the final numbers. 

\begin{table}[h]
    \centering
    \caption{Summary of PlaNet environment and training hyperparameters.}
    \label{tab:planet}
    \begin{tabular}{lll}
        \toprule
        Hyperparameters & 3D Light-Dark\\
        \midrule
        \textbf{Environment} &  &  \\
        Observation size & 64 $\times$ 64 $\times$ 3 \\
        Agent velocity & 0.2 \\
        Action repeat & 2 \\
        Noise amount (image) & 40$\%$ \\
        Occlusion amount (image) & 30 $\times$ 30 \\
        \midrule
        \textbf{Training} &  &  \\
        Number of training steps & 1e6 \\
        Number of training steps per data collection phase & 100 \\
        Environment steps per data collection phase & 200 \\
        Learning rate & 1e-3 \\
        \bottomrule
    \end{tabular}
\end{table}

\end{document}